\documentclass[lettersize,journal]{IEEEtran}
\usepackage{amsmath,amsfonts}
\usepackage{algorithmic}
\usepackage{algorithm}
\pdfoutput=1
\usepackage{array}
\usepackage[caption=false,font=normalsize,labelfont=sf,textfont=sf]{subfig}
\usepackage{textcomp}
\usepackage{stfloats}
\usepackage{url}
\usepackage{verbatim}
\usepackage{graphicx}
\usepackage{arydshln}
\usepackage{cite}
\usepackage{hyperref}
\usepackage{cleveref}
\usepackage{multirow}
\usepackage{graphicx}
\usepackage{booktabs}
\usepackage{amssymb}
\usepackage{subcaption}
\usepackage{amsmath}
\usepackage{xcolor}

\usepackage{soul}
\usepackage{xcolor}
\usepackage{colortbl}
\definecolor{HighlightColor}{RGB}{255,220,220}

\sethlcolor{yellow}

\begin{document}

\newcommand{\snifferemoji}{\includegraphics[height=2.5\fontcharht\font`\B]{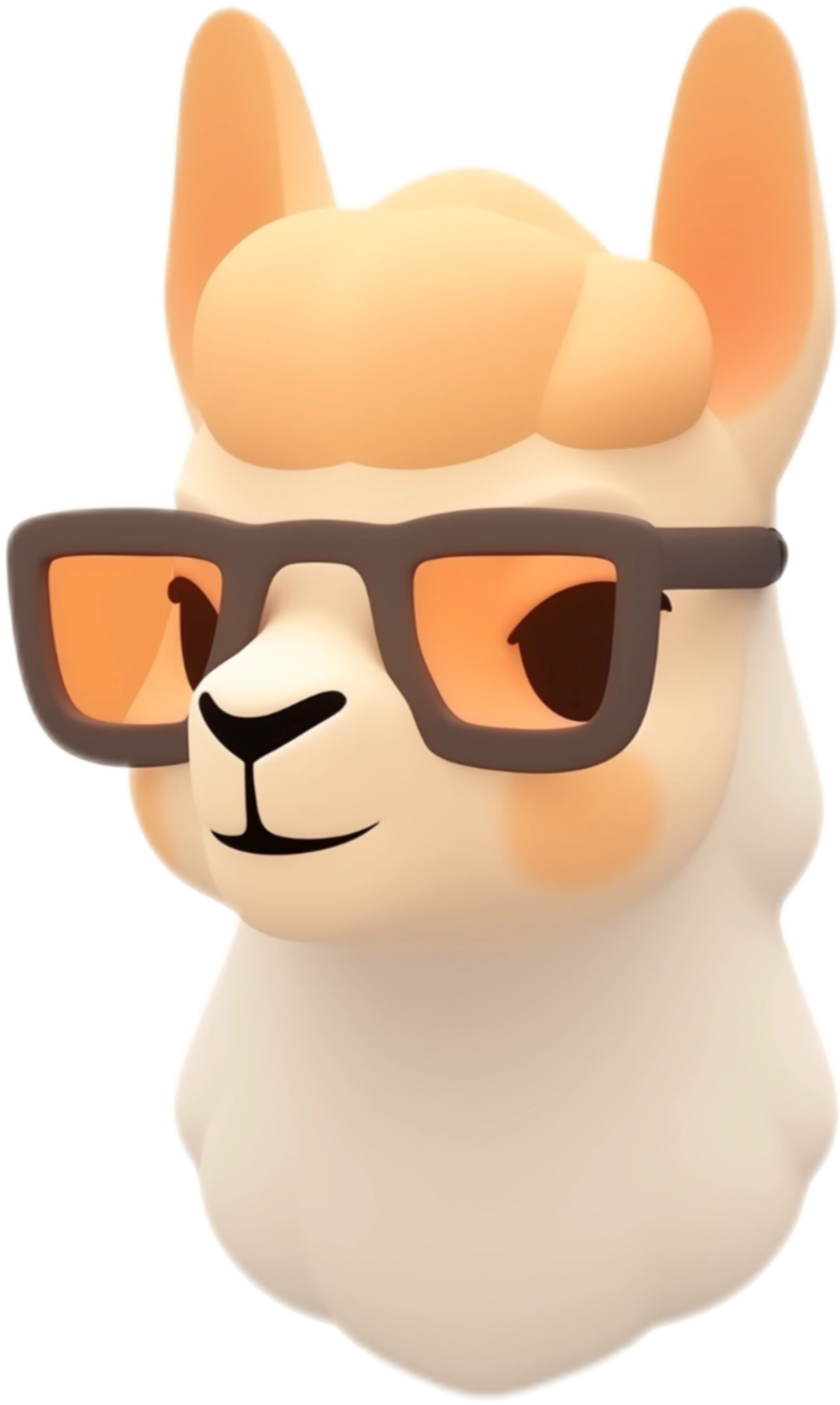}}

\title{
%\snifferemoji{} 
ForgeryGPT: A Multimodal LLM for Interpretable Image Forgery Detection and Localization}

\author{Fanrui Zhang, Jiawei Liu, Jiaying Zhu, Esther Sun,    Dong Li, Qiang Zhang and Zheng-Jun Zha,~\IEEEmembership{Member,~IEEE}
% \renewcommand{\thefootnote}{}
        % <-this % stops a space
\thanks{Jiawei Liu, Fanrui Zhang, Jiaying Zhu, Qiang Zhang, Dong Li and Zheng-Jun Zha are with the Department of Information and Intelligence, University of Science and Technology of China, Hefei, China, 230027 (E-mail: jwliu6@ustc.edu.cn).}

\thanks{Esther Sun is with the Institute for Interdisciplinary Information Sciences, Tsinghua University, Beijing, China, 100190.} 
%E-mail: yy-sun23@mails.tsinghua.edu.cn.}% <-this % stops a space
}

\markboth{IEEE TRANSACTIONS ON IMAGE PROCESSING,~Vol.~xx, No.~xx, January~2026}%
{Liu \MakeLowercase{\textit{et al.}}: A Sample Article Using IEEEtran.cls for IEEE Journals}

% Remember, if you use this, you must call \IEEEpubidadjcol in the second

\maketitle

\begin{figure*}[t]
    \centering
    \includegraphics[width=0.8\textwidth]{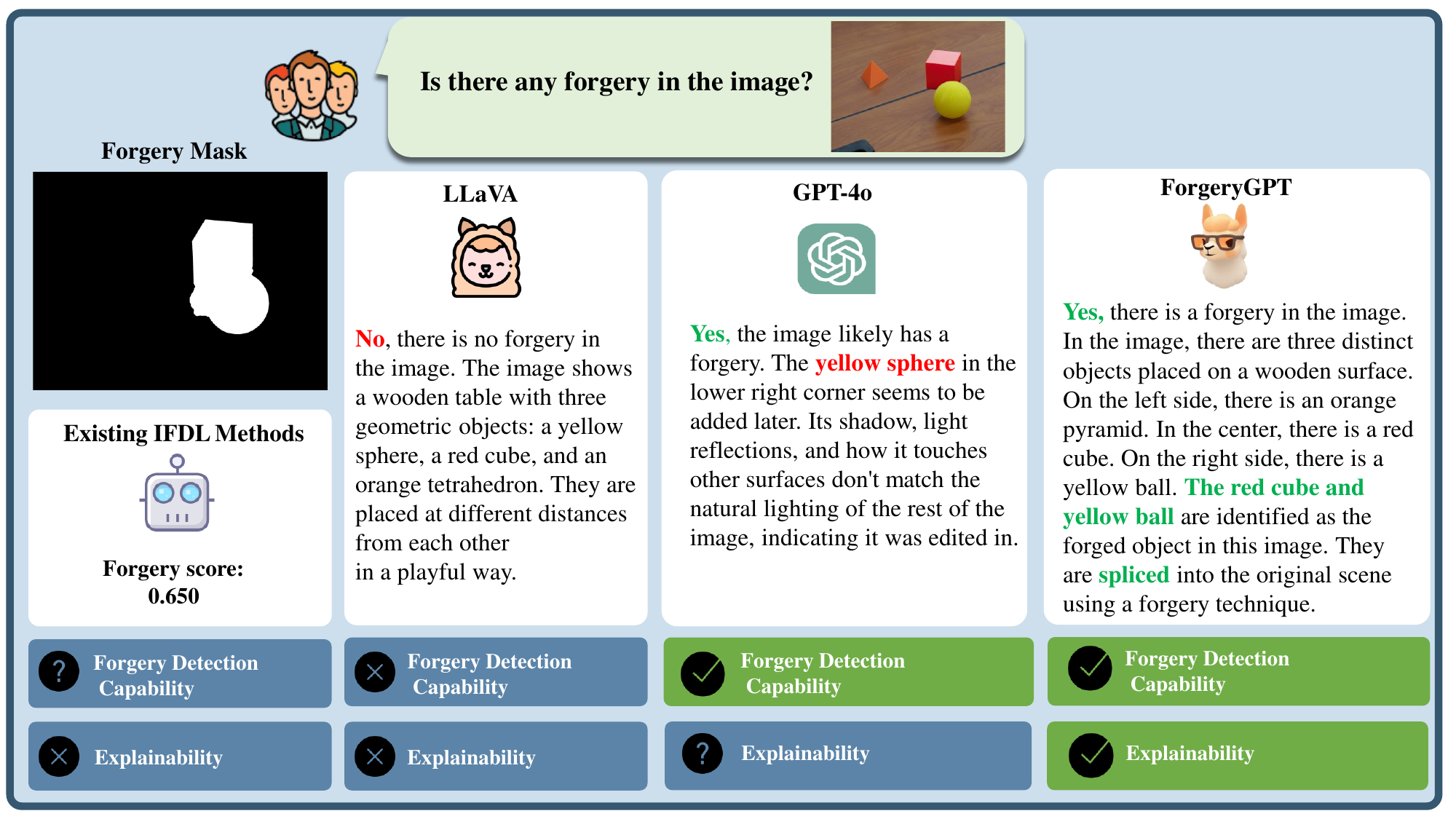}
    \caption{Comparison between our ForgeryGPT and existing approaches.
    ``Forgery Mask'' denotes the ground-truth mask of the manipulated image.
    Current IFDL methods typically report a forgery score and lack interpretability, with detection performance strongly dependent on manually set thresholds.
    Existing MLLMs either do not support forgery detection or fail to provide precise explanations for detected manipulations. In contrast, ForgeryGPT not only correctly detects and localizes forgery traces but also accurately identifies forged objects, classifies forgery types, and delivers a detailed reasoning process.}
    \label{fig:introduction-new}
    \vspace{-3mm}
\end{figure*}

\begin{abstract}
Multimodal Large Language Models (MLLMs), such as GPT4o, have demonstrated strong capabilities in visual reasoning and explanation generation. However, despite these advantages, they face notable challenges in the critical task of Image Forgery Detection and Localization (IFDL), where subtle manipulation traces are often overlooked. Existing IFDL approaches are mostly confined to learning low-level, semantic-agnostic clues, neglecting high-level forensic semantics within forged content, and produce only a single judgment without any reasoning.
% ## 小模型检测依赖设置阈值并且特征利用不够
% 缺少高层次语义，高层次知识挖掘，缺乏解释性
To address these limitations, we propose ForgeryGPT, a novel framework that advances the IFDL task by capturing high-order forensics knowledge correlations from forged images across diverse linguistic feature spaces, while supporting explainable reasoning and interactive dialogue through a customized Large Language Model (LLM).
Specifically, ForgeryGPT enhances traditional LLMs with a Mask-Aware Forgery Extractor, enabling precise excavation of forgery mask information and pixel-level understanding of tampering artifacts. This extractor comprises a Forgery Localization Expert (FL-Expert) and a Mask Encoder. The FL-Expert incorporates an Object-agnostic Forgery Prompt and a Vocabulary-enhanced Vision Encoder to capture multi-scale fine-grained forgery details via cross-modal reasoning, thereby improving tampering localization at the knowledge level. 
% ### 数据训练上的创新, 
For comprehensive training, beyond standard image-text feature alignment, we construct a Mask-Text Alignment Pre-Training dataset using the generated multi-granularity forged images, enabling accurate alignment of forgery masks in the LLM feature space. Subsequently, a Task-Specific Instruction Tuning dataset is employed to refine detection, localization, and dialogue capabilities, yielding robust performance and interpretable inference in forgery-specific scenarios.
Extensive experiments demonstrate that ForgeryGPT substantially outperforms state-of-the-art methods, providing the first step toward integrating IFDL with explainable, multi-turn dialogue capabilities and achieving strong generalization across diverse datasets.
% To further advance forgery detection, we generate a large-scale, multi-granularity pre-training dataset of forged images using simulated forgery techniques. In parallel, we employ a GPT-4-assisted data generation pipeline to produce domain-specific textual descriptions tailored for forgery detection. 
% We also utilize open-source forgery datasets and GPT-4 to create fine-tuning dialogues based on a variety of question templates. These dialogues address tasks such as identifying the presence of forgery, detecting forged objects, and providing detailed explanations for the detected manipulations.
%  紧接着通过我们精心设计的模板构建涉及判断，位置，种类等的SFT数据并通过GPT4o来增强数据集的泛化性。
% ## 大模型通过三阶段的训练，输出可解释性，0泛化能力
%Extensive experimental results across multiple benchmarks demonstrate that ForgeryGPT significantly outperforms state-of-the-art methods, offering the first step towards integrating the IFDL task with explainable, multi-turn dialogue capabilities while also showing remarkable generalization across datasets from diverse domains.
% Extensive experimental results across multiple benchmarks demonstrate that our ForgeryGPT significantly outperforms state-of-the-art methods.
% Through our meticulously designed three-stage instruction tuning with the corresponding constructed data, ForgeryGPT is capable of directly assessing both the presence and locations of forgeries without manually setting thresholds. 
% Additionally, ForgeryGPT supports multi-turn dialogues for outputting detailed explanations and demonstrates remarkable generalization capabilities across datasets from different domains.

\end{abstract}


\section{Introduction}
\IEEEPARstart{T}{he} rapid evolution of media technologies and image editing tools has dramatically increased the prevalence of photo manipulation, resulting in a range of serious societal consequences. These manipulations enable misinformation, forged judicial evidence, and unauthorized watermark removal~\cite{liu2021two,ferreira2016behavior}, and their widespread circulation has even induced public panic and social instability. To address these risks, Image Forgery Detection and Localization (IFDL) aims to verify authenticity and precisely localize tampered regions. Yet, as forgery methods—especially diffusion-based generative techniques—advance rapidly and manipulated media spreads faster, manual verification becomes increasingly infeasible~\cite{li2024image,nichol2021glide,korus2016multi2}. This motivates automated, reliable IFDL systems to safeguard the credibility of visual information in digital ecosystems.

Existing IFDL approaches typically extract low-level forgery cues and learn semantic-agnostic representations through semantic segmentation networks for identifying manipulation at both the image-level and the pixel-level. 
These approaches can be broadly categorized into two types: spatial-domain methods, which focus on identifying low-level visual artifacts such as texture inconsistencies or noise \cite{RGB_Nzhou2018learning,MVSS-Net2021image,zhu2024learning,guillaro2023trufor}; and frequency-domain methods, which exploit high-frequency components to detect imperceptible manipulations \cite{yu2024diffforensics,Objectformerwang2022objectformer}.
%These methods generally fall into two categories: (1) those focused on extracting low-level image features, such as detecting artifacts (\textit{e.g.}, image noise), to capture forgery clues as supplementary image information \cite{RGB_Nzhou2018learning,MVSS-Net2021image,,guillaro2023trufor}, and (2) those aimed at adapting stronger image encoders, like CNN, ViT, or diffusion models, to learn image features from different perspectives \cite{Objectformerwang2022objectformer,yu2024diffforensics,ma2023iml}.
% 定位器的缺点：和多角度特征(类无关的语言特征)
% 到目前为止，最先进的方法通常只关注图像特征的挖掘上，如提取图像的artifacts (e.g., image noise)和适配不同的图像特征编码器。忽略了the exploration of high-level semantic information within forged images ，语言和知识层面对伪造图像进行分析和特征抽取。
Despite these advancements, existing approaches struggle to incorporate high-level forensics semantics and knowledge-guided reasoning into forgery analysis. Their inability to learn higher-order representations limits the detection of more complex manipulations.
%  检测器缺乏推理过程和可解释性
Moreover, most current systems output detection and localization results without providing any interpretable reasoning process. This lack of transparency prevents users from understanding the decision rationale, thereby undermining both trust and accountability.
% 检测器依赖设置阈值来进行判断
Furthermore, these methods typically output a forgery likelihood or confidence score for each sample, depending on manually set thresholds to distinguish genuine from forged samples, which restricts their flexibility and adaptability across varied scenarios.

% 写大模型的部分 \IEEEPARstart{L}{arge}
In recent years, Multimodal Large Language Models (MLLMs) have shown strong performance on vision--language tasks that require visual understanding and natural language reasoning. Models such as GPT-4o~\cite{gpt4} and LLaVA~\cite{llava} demonstrate impressive multimodal alignment and knowledge-driven reasoning. However, directly applying MLLMs to IFDL is challenging: pretraining on web-scale data provides limited forensic expertise, and standard architectures are often insensitive to the subtle artifacts in manipulated regions, hindering precise localization. These limitations motivate domain-specific adaptations for fine-grained and explainable forgery detection and localization.

% 现有大模型和小模型效果都很差
To address these limitations, as illustrated in Figure~\ref{fig:introduction-new}, we introduce ForgeryGPT, a novel framework that substantially advances IFDL by modeling high-order forensic knowledge correlations within forged images across diverse linguistic feature spaces. Simultaneously, it enables explainable reasoning and interactive dialogue through a customized Large Language Model (LLM). This design supports more nuanced forgery analysis, enhances the interpretability, and improves the transparency throughout the detection process.
% ## 对应依赖阈值判断缺陷
Unlike traditional approaches, ForgeryGPT conducts forgery detection without manually defined thresholds and generates accurate localization masks via an integrated Forgery Localization Expert (FL-Expert), providing an end-to-end and explainable solution for IFDL.
Specifically, the framework consists of three components: an Image Encoder, a Mask-Aware Forgery Extractor, and a Large Language Model. Given an input image, ForgeryGPT first predicts a forgery mask and pairs it with textual instructions; these multimodal inputs are embedded into image, mask, and text tokens, interleaved, and fed into the LLM for fine-grained detection and reasoning.
% #思路：提出了ForgeryGPT，三阶段训练(第一阶段用开源的数据集进行模态对齐，第二阶段构建了一个多粒度伪造预训练数据集，并用gpt4来打标和数据生成管道)，（第三阶段构建领域专用的SFT数据，不仅包括是否有异常的判断，还有关于伪造物体是什么，伪造区域在哪里，伪造理由描述等相关对话问题），引入定位小模型（以clip架构为基础，引入了类无关的语言prompt特征，增强的图像特征编码器作为基本图像语义特征的补充），通过多模式指令调整和外部知识的增强，ForgeryGPT展示了显着的性能改进，比原始 MLLM 提高了 40% 以上，并超越了当前最先进的方法。除了准确的判断之外，ForgeryGPT还擅长提供精确且有说服力的解释，定量分析和人工评估证明了这一点。
% 结构流程总述， 对应现有大模型设计问题
% 基本流程
%Given an input image, a forgery mask generated by ForgeryGPT, and text instructions, we tokenize and transform them into token embeddings. The interleaved image, mask, and text tokens are then fed into the LLM to achieve fine-grained forgery understanding. 
% 一个小细节,插入可学习的prompt token
%To enhance this process, we introduce learnable prompt tokens between the image and mask tokens, acting as connectors that bridge modality spaces and facilitate efficient information transfer and integration across modalities.
ForgeryGPT additionally introduces learnable prompt tokens between image and mask tokens to bridge modality gaps and facilitate cross-modal information fusion.
% image encoder
The Image Encoder captures rich spatial details and projects them into a unified embedding space to maintain vision--language feature alignment.
% Mask-Aware Forgery Extractor
To further enhance sensitivity to tampering artifacts and strengthen the robustness, the Mask-Aware Forgery Extractor is developed to capture both pixel-level forgery patterns and mask-level contextual cues. This design significantly boosts the model’s capacity for fine-grained IFDL. 
Within it, the FL-Expert and Mask Encoder serve as the two essential components responsible for forgery mask extraction and semantic refinement. 
The FL-Expert, built upon a CLIP architecture, is augmented with an Object-agnostic Forgery Prompt and a Vocabulary-enhanced Vision Encoder. This dual augmentation allows the module to capture intricate forgery evidence through cross-modal reasoning and facilitate manipulation localization at the semantic knowledge level.
% Object-agnostic Forgery Prompt 
Specifically, the Object-agnostic Forgery Prompt learns category-independent prompts, allowing effective mining of latent forensic knowledge across diverse manipulations.
% Vocabulary-enhanced Vision Encoder
Meanwhile, the Vocabulary-enhanced Vision Encoder extends the CLIP vision encoder with a trainable visual vocabulary network, enriching coarse representations with fine-grained visual cues and improving localization of subtle tampering traces within the CLIP feature space.
% mask encoder
Finally, the Mask Encoder converts forgery masks into tokens aligned with the LLM feature space, enabling seamless fusion with text embeddings for context-grounded detection and localization.

% ###对应现有大模型数据和训练缺陷，我们的三阶段训练和对应的数据集
The training process of ForgeryGPT adopts a comprehensive three-stage optimization paradigm designed to progressively adapt the customized LLM to the intricate requirements of IFDL.
% 第一阶段训练和数据集
1) In the first stage, \textit{Image-Text Alignment Pre-training}, establishes a fundamental correspondence between visual representations and linguistic semantics by utilizing the Image Encoder and a pre-trained LLM on the filtered open-source CC3M \cite{llava} dataset, while only the image-level MLP projector is trainable.
% 第二阶段
2) In the second stage, \textit{Mask-Text Alignment Pre-training}, enhances fine-grained forgery perception and contextual semantics by optimizing the Mask-Aware Forgery Extractor to align region-level features from forgery masks with corresponding text embeddings.
% 总述
To support this, we construct a dedicated pre-training dataset of 76,013 forged image--caption pairs, covering splicing, copy-move, and object removal.
% 先生成伪造图片
These large-scale, multi-granularity forged samples are synthesized via advanced inpainting and localized repainting techniques~\cite{diffusion,nichol2021glide}.
% 再打标
To ensure semantic richness, we employ a carefully designed pipeline where GPT-4o~\cite{gpt4v} generates instruction-oriented annotations with detailed descriptions and auxiliary cues for each manipulated image, enabling comprehensive modeling of forgery characteristics across diverse scenarios.
% 第三阶段
3) In the final stage, \textit{Task-Specific Instruction Tuning}, further enhances pixel-level forgery understanding and multi-turn instruction following. We curate 48,000 dialogue instances from benchmark datasets, including Fantastic-Reality~\cite{kniaz2019point} and CASIAv2~\cite{dong2013casia}, covering forged object identification, tampered-region localization, manipulation-type classification, and forgery reasoning. This stage improves instruction compliance, interpretability, and reasoning coherence.
% ###实验效果巴拉巴拉
The extensive experiments demonstrate that ForgeryGPT achieves notable performance gains over both existing MLLMs (\textit{e.g.}, GPT-4o) and state-of-the-art IFDL approaches. Beyond its superior detection and localization accuracy, ForgeryGPT produces logically consistent, human-interpretable explanations, as evidenced by both quantitative and human evaluation, substantially improving trustworthiness and applicability in real-world forensic contexts.

Our contributions are summarized as follows:
%  首创1，结构2，训练1，数据1，效果，
(1) We pioneer the application of MLLMs to the IFDL domain and propose a novel framework, ForgeryGPT, which not only performs forgery detection and localization without manual threshold adjustment but also supports multi-round dialogue and explainable reasoning.
(2) We enhance conventional LLM architectures by incorporating the Mask-Aware Forgery Extractor, comprising the FL-Expert and the Mask Encoder, that enables precise extraction of forgery masks and facilitates pixel-level comprehension of tampering artifacts.
(3) We develop FL-Expert, a CLIP-based forgery localization module integrating an Object-agnostic Forgery Prompt and a Vocabulary-enhanced Vision Encoder. This design enables the model to generalize across diverse manipulation types by learning semantic prompts independently of specific objects, while enriching visual representations with fine-grained vocabulary cues. 
(4) We introduce a comprehensive three-stage training paradigm for ForgeryGPT, supported by two high-quality datasets, the Mask-Text Alignment Pre-training dataset and the IFDL Task-Specific Instruction Tuning dataset. This progressive strategy establishes vision-language alignment, enhances mask-guided forgery detection, and strengthens instruction-following capabilities, enabling the model to address pixel-level IFDL challenges.
(5) Extensive experiments demonstrate that ForgeryGPT surpasses state-of-the-art methods, pioneering the integration of IFDL with interpretable multi-turn dialogue and exhibiting strong generalization across diverse benchmarks.

\begin{figure*}[t]
    \centering
    \includegraphics[width=0.9\textwidth]{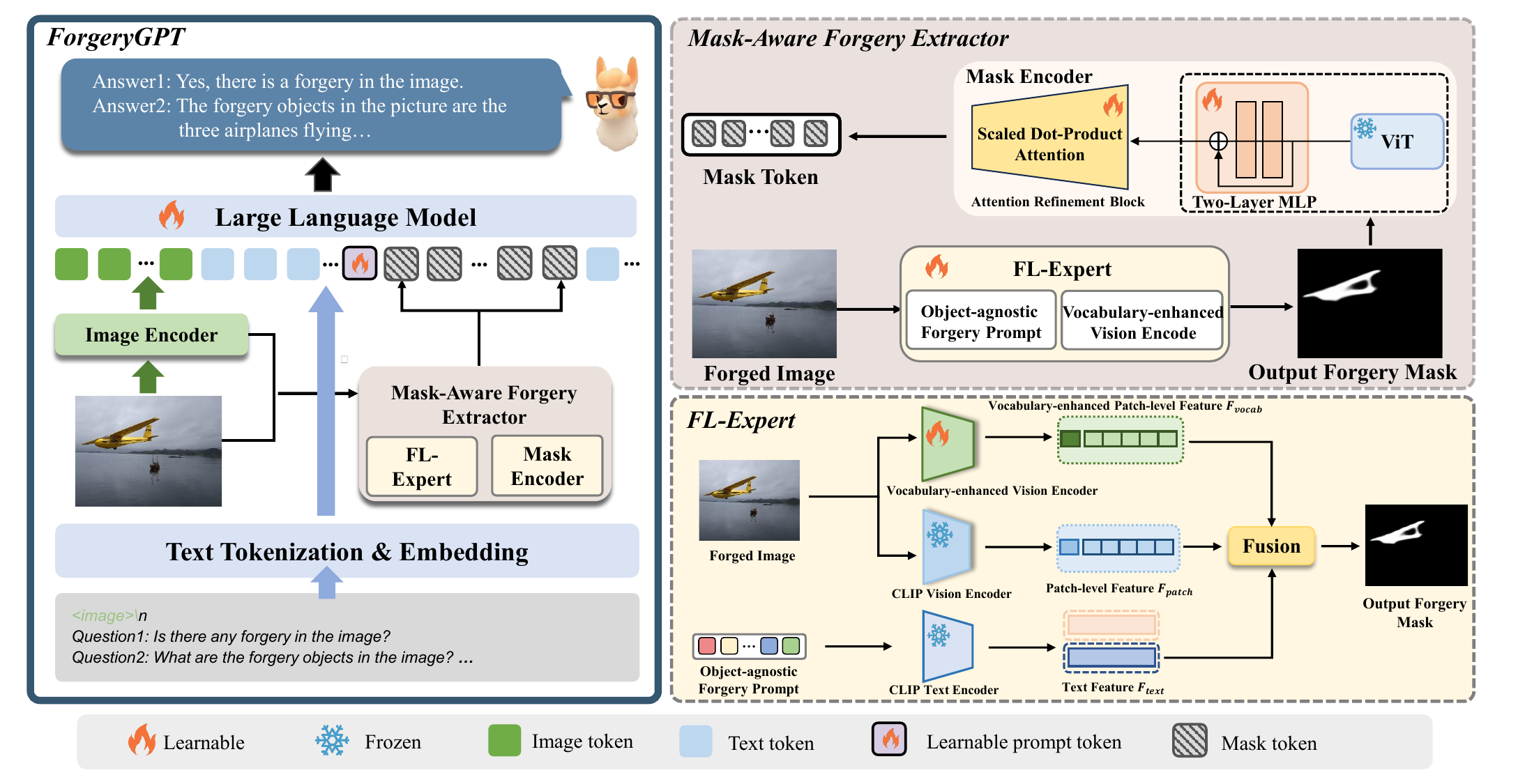}
    \caption{Overview of the proposed ForgeryGPT. The left panel shows the overall architecture, which comprises an Image Encoder, a Mask-Aware Forgery Extractor, and a Large Language Model. The right panel provides a detailed view of the Mask-aware Forgery Extractor and the FL-Expert.}
    \label{fig:framework}
    \vspace{-3mm}
\end{figure*}

\begin{figure*}[t]
    \centering
    \includegraphics[width=0.8\textwidth]{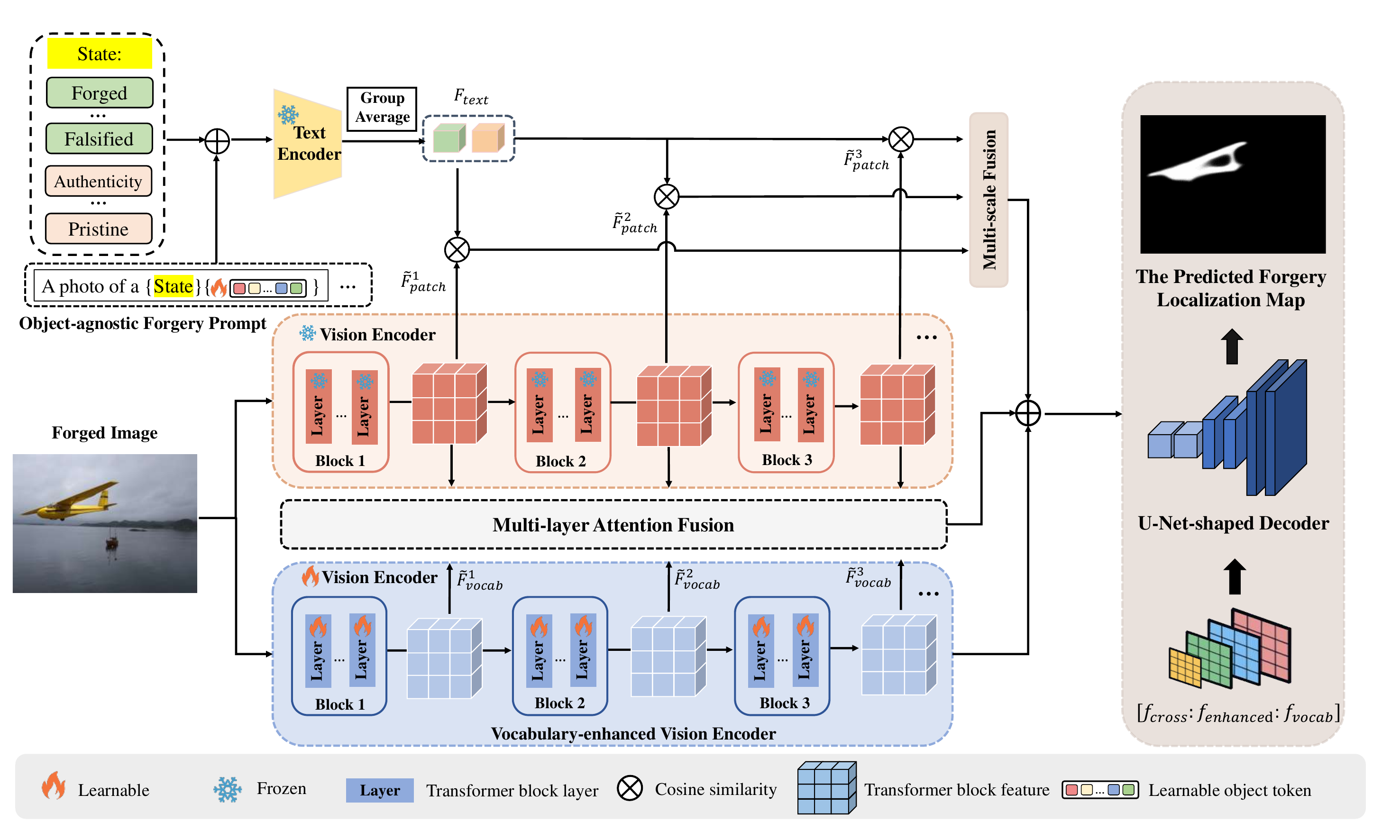}
    \caption{Overview of FL-Expert. It consists of the Object-agnostic Forgery Prompt module, frozen CLIP text and vision encoders, a Vocabulary-enhanced Vision Encoder, Multi-layer Attention Fusion, and a U-Net-shaped Decoder.}
    \label{fig:fl-expert}
    \vspace{-3mm}
\end{figure*}

\section{Related Work}
\subsection{Image Forgery Detection and Localization}
Early research in image forgery localization predominantly targeted specific manipulation traces, such as splicing~\cite{cozzolino2019noiseprint, kniaz2019point}, copy-move~\cite{copymove_wen2016coverage, copymove_wu2018busternet}, and removal~\cite{removal_yang2020spatiotemporal, removal_zhu2018deep}. However, the limited generalization of these methods prompted a shift toward universal forensic frameworks capable of handling diverse forgery types.
To capture subtle manipulation artifacts, recent approaches have exploited multi-domain features. RGB-N~\cite{RGB_Nzhou2018learning} and ManTra-Net~\cite{Mantrawu2019mantra} utilized noise streams and anomaly detection to identify inconsistencies in local features. Subsequent works enhanced representation learning through advanced mechanisms: SPAN~\cite{SPANhu2020span} and ObjectFormer~\cite{Objectformerwang2022objectformer} modeled spatial and frequency dependencies via attention mechanisms, while 
% ERMPC~\cite{li2023edge} and 
TruFor~\cite{guillaro2023trufor} incorporated edge constraints and reliability maps to refine localization boundaries.
Concurrently, holistic frameworks addressing joint detection and localization have emerged. Architectures like MVSS-Net~\cite{MVSS-Net2021image}, CAT-Net~\cite{kwon2021cat}, and PSCC-Net~\cite{PSCCliu2022pscc} leverage multi-scale feature fusion to capture both global semantics and local tampering traces. More recently, HiFi-Net~\cite{HiFi_IFDL} and its extended version~\cite{guo2024language} introduced hierarchical classification and CLIP-guided modules to further boost detection performance.

Despite these advancements, pixel-centric methods often struggle to capture high-level forensic semantics and lack natural language interpretability. This opacity limits user trust in practical applications. To bridge this gap, emerging studies integrate Large Language Models (LLMs) for multimodal reasoning. Zhang \textit{et al.}~\cite{zhang2024common} introduced DD-VQA to facilitate explanation-driven forensics. Similarly, FakeShield~\cite{xu2024fakeshield} and SIDA~\cite{huang2025sidasocialmediaimage} combine visual encoders with LLMs to generate grounded tampering explanations and pixel-level masks. These works underscore a paradigm shift toward more interpretable, vision-language-driven forgery analysis.

% \vspace{-1.0em}
\subsection{Multimodal Large Language Models}
In recent years, the impressive cognitive capabilities of large-scale multimodal models have motivated researchers to explore their potential for visual understanding \cite{blip2}. For example, BLIP-2 \cite{blip2} proposed an architecture in which visual features extracted by an image encoder are fed into an LLM alongside textual prompts. Building upon this approach, models such as LLaVA \cite{llava} and Mini-GPT \cite{minigpt} first focus on aligning image and text representations before performing instruction-guided tuning.
More recently, Qwen2.5-VL~\cite{Qwen2.5-VL} integrated a high-capacity language backbone with a vision encoder to handle a broad spectrum of multimodal tasks, including image captioning, visual question answering, and instruction following. InternVL-2.5~\cite{chen2024internvl} introduced a unified architecture that jointly aligns visual and linguistic modalities through end-to-end training, demonstrating strong capabilities in image-grounded reasoning and visual dialogue. By deeply fusing information from vision and language, these models exhibit robust performance across diverse detection and reasoning tasks. In comparison, DeepSeek-R1~\cite{guo2025deepseek} is primarily a large-scale language model optimized for natural language understanding and generation. Although it achieves strong performance on a range of NLP benchmarks, its functionality is limited to text-only inputs and does not extend to multimodal scenarios involving visual information.

\section{Methodology}
% 先介绍整体模型，再介绍两阶段的训练方案
% 行文结构

\subsection{Architecture Overview}
\label{sec:arch}
% The architecture of ForgeryGPT is illustrated in Figure ~\ref{fig:framework}. ForgeryGPT consists of three core components: an Image Encoder, a Mask-Aware Forgery Extractor, and a Large Language Model (LLM) \cite{vicuna}. 
% For a given input image, a forgery mask generated by FL-Expert, and corresponding text instructions, the model tokenizes and transforms them into token embeddings. These interleaved image, mask, and text tokens are then processed by the LLM, enabling a comprehensive understanding and detection of forgeries. 
% Additionally, to enhance the model’s ability to bridge the gap between visual and textual modalities, learnable prompt tokens are introduced between the image and mask tokens. These tokens serve as connectors, facilitating efficient information transfer and integration across the different modality spaces, improving the overall forgery comprehension capabilities of ForgeryGPT.
As illustrated in Figure~\ref{fig:framework}, the ForgeryGPT architecture integrates an Image Encoder, a Mask-Aware Forgery Extractor (including an FL-Expert), and a Large Language Model (LLM). Given an image and textual instructions, the Image Encoder extracts foundational visual tokens, while the Extractor captures pixel-level tampering artifacts to generate precise mask tokens via the FL-Expert. 
% These interleaved image, mask, and text tokens—fused through learnable prompts to bridge modality gaps—are processed by the LLM. This design enables ForgeryGPT to generate explainable responses covering forgery detection, localization, type identification, and comprehensive manipulation reasoning.

\subsection{Image Encoder}
\label{sec:vetokenizer}
Given the strong representational power of the CLIP pre-trained vision encoder across diverse visual tasks \cite{CLIPradford2021learning}, we adopt the CLIP model based on the Vision Transformer (ViT) architecture. In the context of IFDL, it is crucial not only to align the fundamental visual features of natural images with the LLM’s embedding space but also to preserve the semantic consistency between high-level conceptual representations of forged images and the fine-grained text embeddings within the pre-trained LLM. Accordingly, we employ a two-layer Multi-Layer Perceptron (MLP) for feature alignment. The MLP enables more effective modeling and transformation of complex visual semantics, ensuring consistent correspondence between visual and textual modalities and improving the capability of fine-grained forgery detection.

\subsection{Mask-Aware Forgery Extractor}
\label{sec:veinstructor}
Beyond leveraging coarse-level image features, ForgeryGPT incorporates fine-grained mask-based features that correspond to potential forgery regions. To capture pixel-level tampering artifacts within object areas, we introduce the Mask-Aware Forgery Extractor. This module simultaneously generates precise forgery masks and encodes rich, mask-level visual representations from the input images, substantially improving the model’s sensitivity to manipulated regions.
Concretely, the Mask-Aware Forgery Extractor first generates a visual saliency map highlighting suspected tampered areas through the FL-Expert. Subsequently, the Mask Encoder processes the generated forgery masks to produce dedicated mask tokens aligned with the LLM’s embedding space. These mask tokens, enriched with prior knowledge from the FL-Expert, interact with other learned token embeddings to construct a refined visual-language representation of the forgery regions. Such cross-modal interaction enables the model to perform accurate pixel-level detection and reasoning, thereby enhancing both the granularity and reliability of forgery localization.

% This process not only improves the accuracy of forged image detection but also enhances the model's generalization capability across different forgery techniques.

\subsubsection{FL-Expert}
\label{sec:data_prepare}
Built on CLIP’s strong zero-shot ability to distinguish authentic from manipulated attributes, the FL-Expert is implemented atop CLIP, as shown in Figure~\ref{fig:fl-expert}. The vision and text encoders of CLIP are kept frozen to retain their prior knowledge. On top of this foundation, we introduce two specialized components: the Object-agnostic Forgery Prompt and the Vocabulary-enhanced Vision Encoder, designed to fully harness CLIP’s potential for the IFDL task.

\textbf{Object-agnostic Forgery Prompt.} 
To address the inherent variability and uncertainty of forged objects, we design adaptive object-agnostic prompts that span a range of textual templates. This mechanism allows the model to effectively encode abstract forgery concepts and characterize subtle differences between authentic and manipulated content. Traditional CLIP text prompts, such as ``\textit{A photo of a forged person}'', are primarily constrained to specific object-related semantics, limiting their ability to generalize text embeddings capable of distinguishing forged from authentic content across diverse object categories. 
Consequently, these prompts are insufficient for querying meaningful visual embeddings required for robust forgery reasoning. Moreover, the unknown and highly variable nature of forged objects renders manual prompt design for all potential scenarios infeasible.
To overcome these limitations, we employ learnable text prompt templates capable of representing a wide spectrum of forgery-related semantics. During training, these templates adapt dynamically, producing text embeddings that better discriminate between authentic and forged images. This adaptive mechanism enhances the model’s generalization to diverse forgery contexts. Formally, the Object-agnostic Forgery Prompts are defined as follows:
\begin{equation}
\begin{aligned}
g_p &= [M_1][M_2] \ldots [M_E][{object-p}] \\
g_n &= [N_1][N_2] \ldots [N_E][{object-n}]
\end{aligned}
\end{equation}
Where $g_p$ and $g_n$ correspond to the authentic and forged object-agnostic prompts, respectively. \([M_1][M_2]\ldots[M_E]\) and \([N_1][N_2]\ldots[N_E]\) denote the text embedding vectors for the pre-defined authentic and forged prompt templates, such as ``\textit{A photo of a pristine}'' and ``\textit{A photo of a forged}''.  The terms \([object-p]\) and \([object-n]\) represent the learnable embeddings for the authentic and forged object-agnostic tokens. Together, these components form the comprehensive text representation \(F_{\text{text}}\). 
By dynamically adapting to different forgery types, the Object-agnostic Forgery Prompts enable the model to capture generalized patterns, thereby improving the detection of manipulations across varying domains and object categories. The adaptive prompting strategy enhances the capability of forgery detection by producing more discriminative and generalized text embeddings.
% Moreover, this prompt design is versatile and can be applied to various target domains without any modifications; for instance, there is no need to know the object names or types of forgeries present in the target dataset.
% \\

% \textbf{Multi-scale feature Learning} 

To comprehensively capture forgery cues across multiple spatial scales, the vision encoder is structured into four block layers, with each block layer \(i\) extracting intermediate patch-level representations denoted as \({F}^i_{\text{patch}}\). These representations encode local structural details at different granularities, enabling fine-grained analysis for distinguishing authentic content from manipulated regions. Subsequently, we compute the similarity between the patch-level features \({F}^i_{\text{patch}}\) and the text representation \(F_{\text{text}}\), which jointly encapsulates both genuine and forged semantics.
Since the intermediate patch-level visual representations are not yet aligned with the textual embedding space, direct comparison remains infeasible. To bridge this gap, we introduce additional learnable linear projection layers that transform the intermediate visual features into aligned forms \(\tilde{F}^i_{\text{patch}}\), allowing consistent, authentic, and forged semantics matching with the text embeddings.
The resulting cross-modal reasoning feature \(f_{cross}\), which aggregates forgery-related evidence from both image and text modalities, is derived via multi-scale fusion of these aligned representations:
\begin{equation}
\begin{aligned}
f_{cross} = \text{Mean} \left( \sum_{i=1}^{4} \text{softmax} \left( \tilde{F}_\text{patch}^i F_\text{text}^T \right) \right)
\end{aligned}
\end{equation}

\textbf{Vocabulary-enhanced Vision Encoder.} 
To mitigate the limitation of CLIP in recognizing locally forged images, we develop a Vocabulary-enhanced Vision Encoder that incorporates a trainable visual vocabulary network sharing the same ViT-based architecture \cite{vit} as CLIP. This design enables the FL-Expert to seamlessly integrate newly learned, domain-specific vocabulary knowledge with CLIP’s existing open-world recognition capability. By emphasizing the specific visual characteristics of tampered content, this combination improves the model’s precision and robustness in forged image identification.
Following the same processing pipeline, intermediate patch-level representations are extracted and passed through a linear transformation to yield visual vocabulary-encoded features \(\tilde{F}^i_{\text{vocab}}\) at each block layer $i$. To fuse the multi-scale semantic features derived from both the original CLIP encoder and the vocabulary-enhanced encoder, we introduce a Multi-layer Attention Fusion mechanism. This component computes attention between the patch-level features and the visual vocabulary-encoded features across multiple transformer layers of the two vision encoders. For each transformer block layer $i$, the attention weights $\alpha^i$ are computed as follows:
\begin{equation}
\begin{aligned}
\alpha^i &= \text{Softmax}\left({{{\tilde{F}}_{patch}^{i}}{\{\tilde{F}_{vocab}^{i}\}}^T}/{\sqrt{d_e}}\right) 
\end{aligned}
\end{equation}
where ${\tilde{F}}_{patch}^{i}$ and ${\{\tilde{F}_{vocab}^{i}\}}$ serve as the query and key vectors, respectively, and $d_e$ denotes the feature dimension. The attention operation models fine-grained semantic interactions between patch-level features and visual vocabulary-encoded representations. The resulting attention output is computed as:
\begin{equation}
\begin{aligned}
f_{enhaned} = \text{Mean} \left( \sum_{i=1}^{4} \alpha^i \cdot \tilde{F}_{\text{vocab}}^{i} \right)
\end{aligned}
\end{equation}
where ${\tilde{F}_{vocab}^{i}}$ denotes the value vector, $f_{enhanced}$ represents the vocabulary-enhanced reasoning feature. This output complements other extracted features, enriching the model’s understanding of tampered content by providing more detailed and context-specific visual semantics.
By jointly modeling macro-level and micro-level features, the proposed structure is able to uncover the fine-grained characteristics of manipulated images, thereby enhancing the robustness of forgery pattern learning. Finally, the three feature representations are concatenated and passed into a U-Net-shaped decoder \cite{yu2024diffforensics}, consisting of four convolutional layers, to produce the forgery localization mask that delineates the manipulated regions:
\begin{equation}
\begin{aligned}
{G}_{out} = \text{Decoder} \left(  f_{cross}: f_{enhanced}: f_{vocab} \right)
\end{aligned}
\end{equation}
where ${Decoder}$ denotes the U-Net-shaped decoder, $:$ indicates the concatenate operation. $f_{vocab}$ is the final output from the Vocabulary-enhanced Vision Encoder. The resulting ${G}_{out}$ represents the predicted forgery localization map.
% ${G}_{out}$ is the predicted forgery localization map. 
Equipped with the Object-agnostic Forgery Prompt and the Vocabulary-enhanced Vision Encoder, FL-Expert effectively combines multi-scale semantic reasoning, cross-modal fusion, and vocabulary-driven enhancement,
enabling accurate identification and localization of forged content.

\subsubsection{Mask Encoder}
The well-trained FL-Expert produces forgery masks that encapsulate rich prior knowledge essential for IFDL. To effectively transfer this knowledge for the LLM, we introduce a dedicated Mask Encoder, designed to transform the generated forgery mask into mask tokens that the LLM can understand. 
% 在Image Encoder基础上加的block
As illustrated in Figure~\ref{fig:framework}, the Mask Encoder adopts an architecture analogous to the Image Encoder, augmented with an attention refinement block. It generates semantically enriched mask tokens based on the prior knowledge embedded in the FL-Expert while simultaneously compressing the pixel-level forgery details contained within the mask, facilitating compact yet informative integration within the LLM’s embedding space.
% 看看这句话要不要
% Through expert-driven visual-language extraction and feature compression, ForgeryGPT not only achieves accurate forgery detection using high-quality forgery maps generated by the visual expert but also maintains comparable performance when the visual expert fails.
Through expert-guided visual-language abstraction and feature compression, the Mask-Aware Forgery Extractor achieves accurate manipulation detection based on high-quality forgery localization maps from the FL-Expert, while maintaining comparable performance even when the FL-Expert’s predictions are less reliable.

\subsection{Loss Function}

The overall loss of ForgeryGPT consists of two components: a detection loss and a localization loss. These components jointly supervise the model to produce image-level classification labels while simultaneously generating pixel-wise manipulation masks, ensuring both reliable forgery detection and precise localization.

\textbf{Forgery Detection Loss.}  
For the binary image-level forgery detection task, we employ the standard cross-entropy loss, denoted as $\mathcal{L}_{cls}$. Let $p = f_{\theta}^{cls}(x)$ represent the predicted probability that image $x$ is forged, and $y \in \{0, 1\}$ denote the corresponding ground-truth label. The cross-entropy loss is formulated as:
\begin{equation}
\mathcal{L}_{cls} = -[y \log(p) + (1 - y) \log(1 - p)]
\end{equation}
This loss effectively guides the model to distinguish authentic images from manipulated ones.

\textbf{Forgery Localization Loss.}  
For the pixel-level forgery localization task, we adopt the Dice loss~\cite{zhu2024learning}, denoted as $\mathcal{L}_{loc}$, which measures the similarity between the predicted binary mask $G_{out}$ and the ground truth mask $Y$. Dice loss is particularly suited for imbalanced segmentation scenarios and is computed as:
\begin{equation}
\mathcal{L}_{loc} = 1 - \frac{2 \cdot \sum_{i=1}^{N} G_{out}^{(i)} Y^{(i)} + \epsilon}{\sum_{i=1}^{N} G_{out}^{(i)} + \sum_{i=1}^{N} Y^{(i)} + \epsilon}
\end{equation}
where $N$ is the total number of pixels, $G_{out}^{(i)}$ and $Y^{(i)}$ correspond to the predicted and ground-truth mask values at pixel $i$, and $\epsilon$ is a small constant to prevent division by zero.
 
The total training objective is formulated as a weighted sum of the detection and localization losses:
\begin{equation}
\mathcal{L} = \lambda_1 \cdot \mathcal{L}_{cls} + \lambda_2 \cdot \mathcal{L}_{loc}
\end{equation}
Where $\lambda_1$ and $\lambda_2$ are hyperparameters that balance the contributions of each loss component.
By jointly optimizing $\mathcal{L}_{cls}$ and $\mathcal{L}_{loc}$, ForgeryGPT is encouraged to perform robust image-level classification while simultaneously generating precise pixel-level masks, enhancing its capability to detect diverse manipulations and reduce false positives on authentic regions.

\begin{figure*}[t]
    \centering
    \includegraphics[width=1\textwidth]{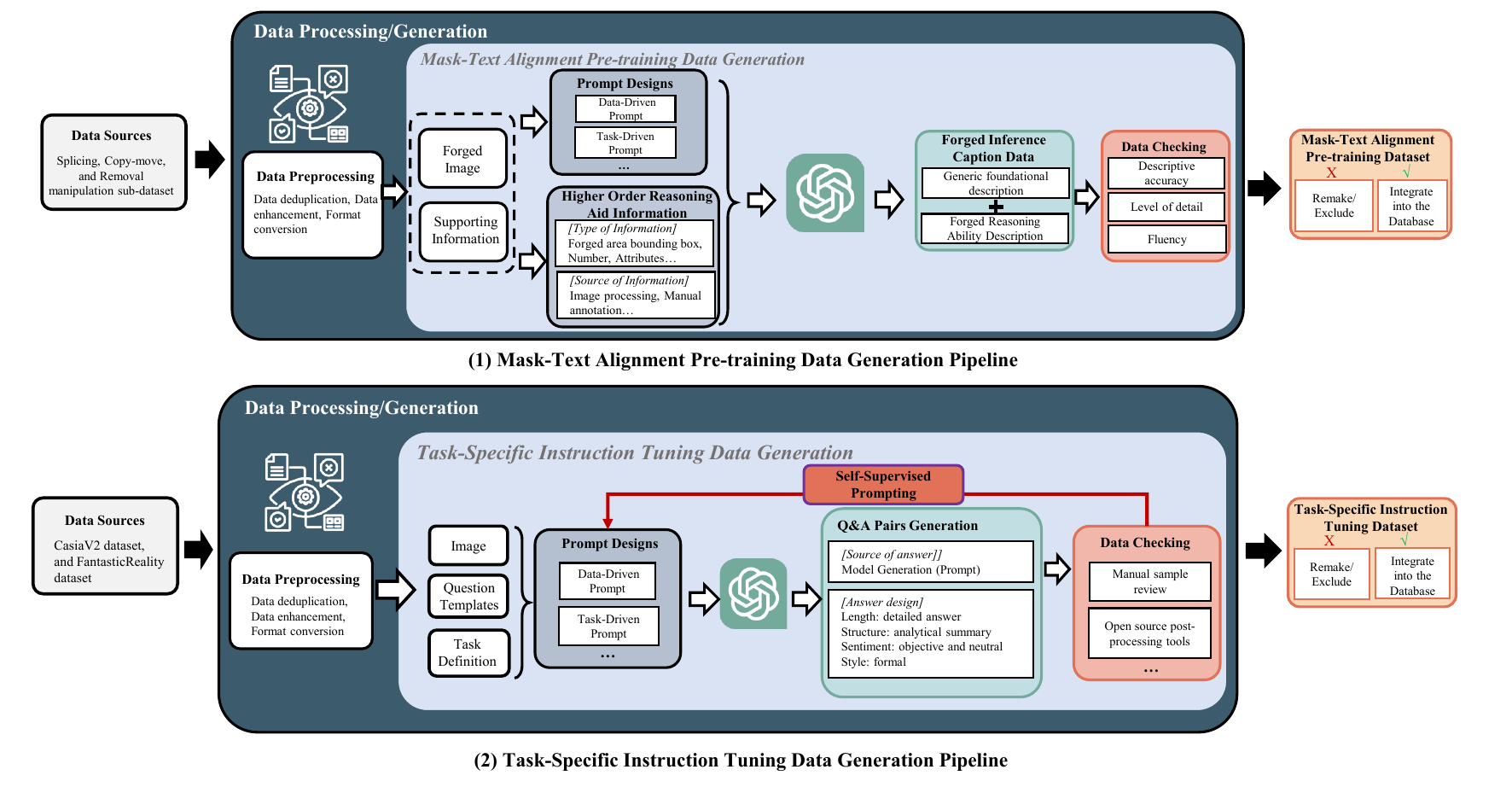}
    \caption{Two training data generation pipelines: one for Mask-Text Alignment Pre-training, which creates caption data from forgery datasets, and another for Task-Specific Instruction Tuning, generating Q\&A pairs for the IFDL task.}
    \label{fig:pipeline}
\end{figure*}

\begin{figure}[!t]
    \centering
    \includegraphics[width=0.9\linewidth]{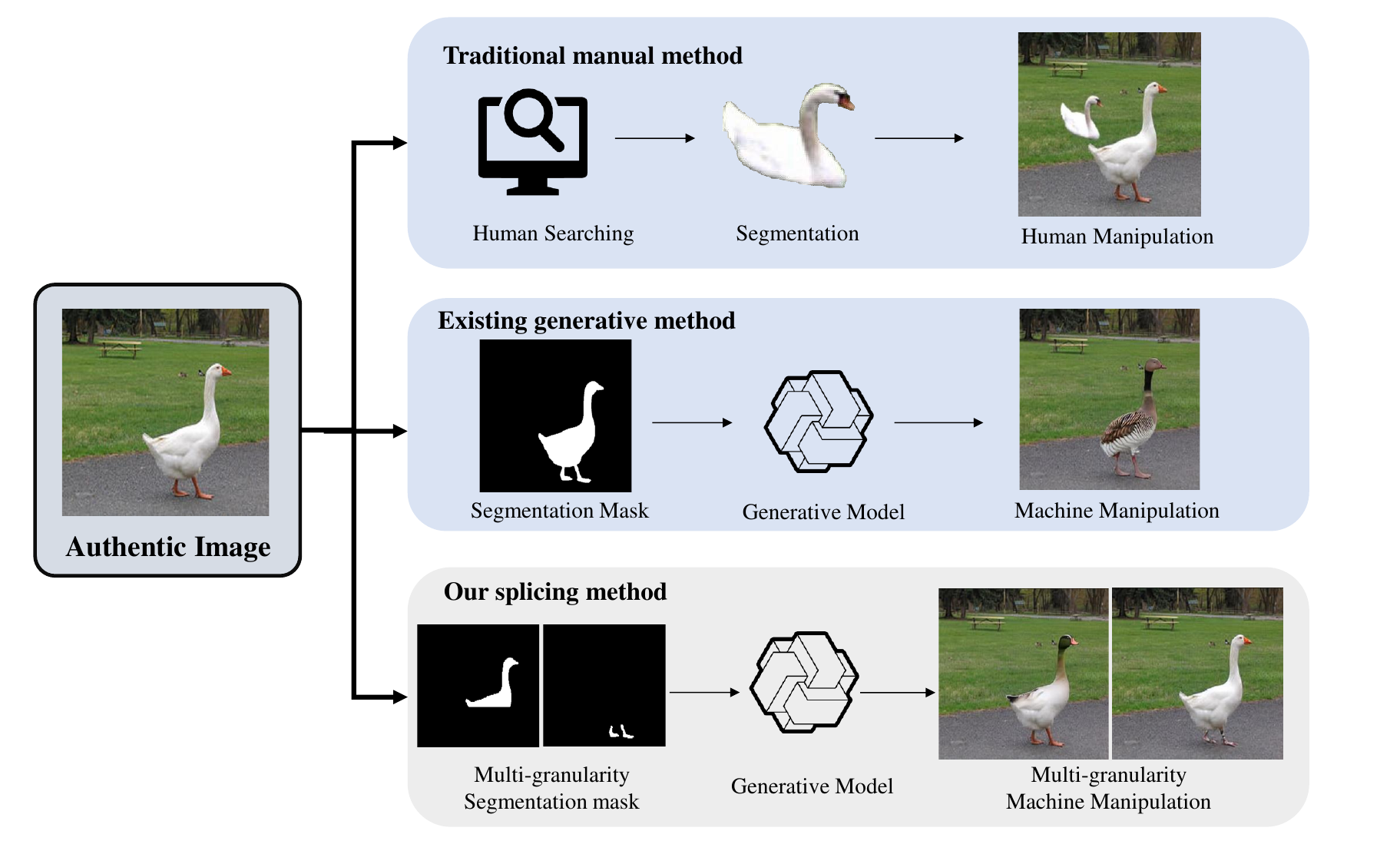}
    \caption{Our splicing method for constructing a multi-granularity splicing manipulation sub-dataset, which uses detailed, multi-granularity segmentation masks to improve image manipulation, compared to the traditional manual method and existing generative techniques.}
    \vspace{-5mm}
    \label{fig:dataset}
\end{figure}

\subsection{Instruction Tuning}
The training procedure of ForgeryGPT involves three stages to progressively adapt the model to the intricacies of the IFDL task, facilitating robust vision-language alignment, precise mask-guided forgery understanding, and specialized instruction tuning.

\subsubsection{Stage 1: Image-Text Alignment Pre-training} 
The first stage aligns global image representations with textual embeddings by training only the image-level MLP projector with a frozen pre-trained LLM. Following LLaVA~\cite{llava}, we use the filtered CC3M dataset and keep both the CLIP vision encoder and LLM weights fixed. This pre-training establishes a strong natural-image alignment foundation for subsequent mask-guided learning and fine-grained forgery reasoning.

\subsubsection{Stage 2: Mask-Text Alignment Pre-training} 
% 阶段目标
Existing open-source LLMs typically learn image-level projectors only from natural images, which is insufficient for IFDL, where manipulation cues are subtle and localized. Building on Stage~1, we introduce a Mask-Aware Forgery Extractor to align region-level features extracted from forgery masks with the LLM embedding space, enabling mask-conditioned interpretation and language-friendly forgery evidence representation. To facilitate this transition from general vision--language understanding to IFDL-specific reasoning, we construct a Mask-Text Alignment pre-training dataset with 76,013 forged image--caption pairs covering three manipulation types: splicing, copy-move, and removal.
% 阶段构建方案
For copy-move, we synthesize samples on MS COCO~\cite{MS-COCOlin2014microsoft} using the method in~\cite{BusterNetwu2018busternet}. For removal, we randomly delete annotated regions and inpaint them with a state-of-the-art model~\cite{inpaintingM2020recurrent}. To better approximate real-world conditions, we additionally apply random Gaussian noise and JPEG compression.

For splicing, we propose a multi-granularity generation strategy to avoid the artifacts of manual pipelines~\cite{guan2019nist} and the oversized tampered regions produced by recent generative methods~\cite{guillaro2023trufor, jia2023autosplice, huang2025sidasocialmediaimage}. As shown in Figure~\ref{fig:dataset}, we generate localized, subtle splicing traces via fine-grained repainting guided by generative models. Using MS COCO annotations, we build multi-granularity local masks via random segmentation and SAM text-prompt segmentation~\cite{kirillov2023segment}: SAM isolates fine parts for distinctive categories (\textit{e.g.}, tails/ears of dogs), while random segmentation yields realistic local masks for less distinctive ones (\textit{e.g.}, bottles, clocks). This adaptive design produces diverse tampering patterns and improves generalization to fine-grained forgery scenarios.

% \begin{figure*}
%     \centering
%     \includegraphics[width=0.8\textwidth]{figs/SFT-case.pdf}
%     \caption{An example from the dataset to illustrate the Task-Specific Instruction Tuning data.}
%     \label{fig:SFT-case}
% \end{figure*}

\begin{table*}
\centering
\vspace{-1em}
\caption{Pixel-level F1 and AUC performance of image forgery localization. Method with * uses the pre-training model of the original paper.}
\resizebox{1\linewidth}{!}{
\begin{tabular}{ccccccccccccccccccc}
\toprule
\multirow{3}{*}{\textbf{Methods}} & \multicolumn{12}{c}{\textbf{Editing}}                                                                                                                                                                                                                                                       & \multicolumn{4}{c}{\textbf{DGM}}                                                                                   & \multicolumn{2}{c}{\multirow{2}{*}{\textbf{Average}}}        \\ 
\cmidrule(lr){2-13} \cmidrule(lr){14-17}
                         & \multicolumn{2}{c}{CASIA1.0+} & \multicolumn{2}{c}{Columbia}    & \multicolumn{2}{c}{NIST16}                      & \multicolumn{2}{c}{IMD2020}                         & \multicolumn{2}{c}{DSO-1}                           & \multicolumn{2}{c}{Korus}                           & \multicolumn{2}{c}{AutoSplice}                      & \multicolumn{2}{c}{OpenForensics}                   & \multicolumn{2}{c}{}                                \\ 
\cmidrule(lr){2-19}
                         & F1            & AUC          & F1   & AUC                      & F1                       & AUC                  & F1                       & AUC                      & F1                       & AUC                      & F1                       & AUC                      & F1                       & AUC                      & F1                       & AUC                      & F1 & AUC \\ 
\hline
ManTraNet* \cite{Mantrawu2019mantra}   & .136 & .612 & .357 & .767 & .160 & .741 & .180 & .785 & .089 & .687 & .104 & .681 & .192 & .622 & .043 & .678 & .158 & .697 \\
HP-FCN \cite{li2019localization}       & .132 & .772 & .050 & .549 & .071 & .690 & .029 & .665 & .013 & .545 & .076 & .663 & .029 & .556 & .027 & .650 & .053 & .636 \\
GSR-Net \cite{zhou2020generate}        & .244 & .821 & .340 & .836 & .221 & .746 & .102 & .788 & .055 & .697 & .060 & .632 & .047 & .722 & .025 & .683 & .137 & .741 \\
SPAN \cite{SPANhu2020span}             & .088 & .533 & .213 & .597 & .116 & .648 & .108 & .671 & .059 & .564 & .070 & .575 & .047 & .572 & .014 & .682 & .089 & .605 \\
MVSS-Net* \cite{MVSS-Net2021image}     & .451 & .845 & .665 & .818 & .292 & .791 & .264 & .817 & .271 & .732 & .095 & .641 & .333 & .839 & .056 & .702 & .303 & .773 \\
CAT-Net \cite{kwon2021cat}             & .394 & .788 & .854 & .826 & .336 & .780 & .295 & .823 & .135 & .713 & .149 & .672 & .185 & .796 & .003 & .552 & .294 & .744 \\
SATL-Net \cite{zhuo2022self}           & .064 & .545 & .677 & .872 & .175 & .655 & .142 & .671 & .084 & .575 & .039 & .577 & .103 & .590 & .019 & .544 & .163 & .629 \\
PSCCNet \cite{PSCCliu2022pscc}         & .355 & .738 & .672 & .881 & .238 & .740 & .295 & .800 & .318 & .721 & .156 & .623 & .150 & .784 & .065 & .610 & .281 & .737 \\
HiFi-Net \cite{HiFi_IFDL}              & .092 & .642 & .382 & .608 & .172 & .685 & .178 & .675 & .304 & .700 & .088 & .607 & \textbf{.613} & .831 & \textbf{.149} & .676 & .247 & .678 \\
\rowcolor{HighlightColor}
{HiFi-Net++ \cite{guo2024language}} &
{.543} & {.870} & 
{.616} & {.920} & 
{.286} & {.801} & 
{.424} & {.793} & 
{.293} & {.681} & 
{.155} & {.696} & 
{.444} & {.808} & 
{.117} & {.703} & 
{.360} & {.784} \\
DiffForensics \cite{yu2024diffforensics} & .517 & .868 & \textbf{.912} & \textbf{.931} & .415 & .828 & .511 & .911 & .485 & .874 & .257 & .721 & .507 & \textbf{.940} & .122 & \textbf{.820} & .466 & .862 \\
{FAKESHIELD \cite{PSCCliu2022pscc}} & 
{.387} & {.774} & 
{.726} & {.899} & 
{.443} & {.845} & 
{.375} & {.818} & 
{.409} & {.797} & 
{.156} & {.667} & 
{.211} & {.805} & 
{.101} & {.590} & 
{.351} & {.774} \\

{SIDA \cite{huang2025sidasocialmediaimage}} & 
{.298} & {.696} & 
{.737} & {.888} & 
{.385} & {.787} & 
{.351} & {.805} & 
{.451} & {.810} & 
{.127} & {.682} & 
{.110} & {.594} & 
{.098} & {.580} & 
{.320} & {.730} \\
\hline
Ours                                   & \textbf{.569} & \textbf{.876} & .773 & .911 & \textbf{.549} & \textbf{.883} & \textbf{.530} & \textbf{.914} & \textbf{.506} & \textbf{.895} & \textbf{.258} & \textbf{.737} & .570 & .877 & .138 & .701 & \textbf{.536} & \textbf{.870} \\
\bottomrule
\end{tabular} 
}
\label{tab:localization}
\vspace{-1em}
\end{table*}

After synthesizing the three forgery types (Fig.~\ref{fig:pipeline}(1)), we use GPT-4o~\cite{gpt4v} to generate instruction-oriented image--caption pairs. We augment prompts with auxiliary metadata (e.g., tampered-region bounding boxes and entity names) to elicit forgery-aware descriptions tailored to each image. The generated captions are then post-processed and validated following~\cite{woodpecker} to ensure reliability and semantic fidelity, yielding captions that accurately capture forgery-specific details for fine-grained feature learning.
% An example of the Mask–Text Alignment Pre-training data is presented in Figure~\ref{fig:pretrain_case}, highlighting that the proposed pipeline yields detailed and high-quality descriptions.

\subsubsection{Stage 3: Task-Specific Instruction Tuning} 

This stage further strengthens ForgeryGPT’s pixel-level forgery understanding while improving instruction following. We freeze the Image Encoder and fine-tune the Mask-Aware Forgery Extractor together with the LLM to enhance responsiveness to detailed prompts and complex forgery reasoning.

As shown in Fig.~\ref{fig:pipeline}(2), we build a task-specific instruction-tuning dataset of 48,000 multi-turn dialogues using forged images from Fantastic-Reality~\cite{kniaz2019point} and CASIAv2~\cite{dong2013casia}. We define core forensic tasks—authenticity judgment, manipulated-region localization, forgery-type identification, and manipulation explanation—and use GPT-4o to generate dialogues from structured question templates. After post-processing and manual validation, the final set contains 48,000 GPT-assisted dialogues for forged cases, paired with an equal number of dialogues for real images (e.g., ``No, there is no forgery information in this image'').

%\\

\section{Experiments}
\subsection{Experimental Settings}
% \
% \noindent{\bf Dataset.} 
\subsubsection{Dataset}
Following the experimental protocol established in \cite{yu2024diffforensics}, we utilize the CASIAv2 \cite{dong2013casia} and Fantastic-Reality \cite{kniaz2019point} datasets, along with their associated detailed forgery annotations, as the foundational data sources for the final stage of Task-Specific Instruction Tuning. To comprehensively evaluate the robustness and generalization capability of our approach, we conduct extensive experiments on a wide range of benchmark manipulation datasets, including CASIA1.0+ \cite{dong2013casia}, Columbia \cite{hsu2006columbia}, NIST16 \cite{guan2019nist}, IMD2020 \cite{novozamsky2020imd2020}, DSO-1 \cite{dso}, and Korus \cite{korus2016multi}. In addition, we extend our evaluation to include cutting-edge datasets synthesized by advanced deep generative models (DGMs), specifically AutoSplicing \cite{jia2023autosplice} and OpenForensics \cite{le2021openforensics}.

% The effectiveness of our method, ForgeryGPT, was systematically compared with state-of-the-art (SOTA) techniques on our custom-developed multi-granularity forgery dataset, thereby providing a comprehensive analysis of its performance across varied forgery contexts.

\subsubsection{Evaluation Metrics} 
To evaluate forgery localization performance, in line with prior studies \cite{guillaro2023trufor, yu2024diffforensics}, we employ pixel-level metrics, including Area Under Curve (AUC) and F1-score, to quantify the accuracy of predicted manipulation masks. For image-level detection performance, ForgeryGPT operates without requiring manually set thresholds; therefore, we directly report image-level accuracy (ACC) and F1-score. For comparative methods that rely on thresholding, a default value of 0.5 is applied to ensure fair and consistent evaluation.

\subsubsection{Implementation Details} 
All experiments are conducted using the PyTorch framework equipped with four NVIDIA Tesla A100 GPUs. The specific configurations for each of the three training stages are detailed as follows:

Stage 1: Image-Text Alignment Pre-training. In this stage, we only train the image-level MLP projector within the Image Encoder using the Adam optimizer. The learning rate is linearly decayed from 1e-4 to 1e-6 over the course of training. This stage is conducted for one epoch with a batch size of 8.

Stage 2: Mask-Text Alignment Pre-training. This stage incorporates the Mask-Aware Forgery Extractor into the training pipeline. Input images are resized to $336 \times 336$ pixels. The backbone of the FL-Expert is the CLIP ViT-B/16 \cite{CLIPradford2021learning}. Object-agnostic learnable word embeddings, corresponding to the tokens $[object-p]$ and $[object-n]$, are set to a dimensionality of 12. Multi-scale features are extracted from transformer blocks \{4, 8, 12, 16\} within the CLIP ViT architecture. 
This stage is trained for five epochs with a batch size of 8.

Stage 3: Task-Specific Instruction Tuning. For this stage, Vicuna-7B \cite{vicuna} serves as the inference LLM. The Mask-Aware Forgery Extractor and the LLM are jointly fine-tuned. The learning rate is set to 2e-5, and the batch size is increased to 32. Optimization is performed using AdamW with a cosine annealing scheduler over a single epoch. The maximum input sequence length for the LLM is configured to 2048 tokens.

\begin{table*}
\centering
\vspace{-1em}
\caption{{Image-level ACC and F1-score performance of image forgery detection.}}
\resizebox{0.7\textwidth}{!}{
\begin{tabular}{ccccccccccc}
\toprule
\multirow{2}{*}{{Methods}} & \multicolumn{2}{c}{{CASIA1.0+}} & \multicolumn{2}{c}{{Columbia}} & \multicolumn{2}{c}{{IMD2020}} & \multicolumn{2}{c}{{AutoSplice}} & \multicolumn{2}{c}{{Average}} \\ 
\cmidrule(lr){2-3} \cmidrule(lr){4-5} \cmidrule(lr){6-7} \cmidrule(lr){8-9} \cmidrule(lr){10-11}
& {ACC} & {F1} & {ACC} & {F1} & {ACC} & {F1} & {ACC} & {F1} & {ACC} & {F1} \\ 
\hline
{H-LSTM \cite{bappy2019hybrid}} & {.535} & {.451} & {.496} & {.411} & {.829} & {.551} & {.614} & {.613} & {.619} & {.507} \\
{ManTra-Net \cite{Mantrawu2019mantra}} & {.535} & {.524} & {.496} & {.481} & {.830} & {.666} & {.614} & {.601} & {.619} & {.568} \\
{GSR-Net \cite{zhou2020generate}} & {.595} & {.581} & {.540} & {.511} & {.671} & {.644} & {.568} & {.509} & {.594} & {.561} \\
{MVSS-Net \cite{MVSS-Net2021image}} & {.791} & {.733} & {.664} & {.589} & {.799} & {.717} & {\textbf{.809}} & {.711} & {.766} & {.688} \\
{CAT-Net \cite{kwon2021cat}} & {.671} & {.655} & {.755} & {.701} & {.785} & {.598} & {.699} & {.590} & {.728} & {.636} \\
{SATL-Net \cite{zhuo2022self}} & {.459} & {.433} & {.744} & {.705} & {.667} & {.651} & {.463} & {.392} & {.583} & {.545} \\
{PSCCNet \cite{PSCCliu2022pscc}} & {\textbf{.992}} & {.614} & {.606} & {.636} & {.821} & {.688} & {.733} & {.642} & {.788} & {.645} \\
{HiFi-Net \cite{HiFi_IFDL}} & {.632} & {.687} & {.532} & {.644} & {\textbf{.826}} & {\textbf{.834}} & {.618} & {.606} & {.652} & {.693} \\

{HiFi-Net++ \cite{guo2024language}} & 
{.822} & {\textbf{.812}} & 
{.815} & {.794} & 
{.781} & {.766} & 
{.737} & {.708} & 
{.789} & {.770} \\
DiffForensics \cite{yu2024diffforensics} & .741 & {.681} & .895 & {.903} & .749 & {.841} & .696 & {.679} & .770 & {.776} \\
{FAKESHIELD \cite{xu2024fakeshield}} & 
{.703} & {.689} & 
{.859} & {.845} & 
{.645} & {.633} & 
{.661} & {.659} & 
{.767} & {.757} \\
{SIDA \cite{huang2025sidasocialmediaimage}} & 
{.736} & {.727} & 
{.738} & {.751} & 
{.730} & {.711} & 
{.603} & {.601} & 
{.702} & {.698} \\
\hline
Ours & .805 & {.794} & \textbf{.903} & {\textbf{.900}} & .727 & {.709} & .780 & {\textbf{.760}} & \textbf{.804} & {\textbf{.791}} \\
\bottomrule
\end{tabular}
}
\label{tab:detection}
\vspace{-1em}
\end{table*}

\begin{table*}[ht]
    \centering
    \caption{{Ablation results on CASIA1.0+, Columbia,
 IMD2020 and AutoSplice datasets using different variants of our proposed ForgeryGPT on the image forgery detection task.}}
    \resizebox{0.7\linewidth}{!}{
   \begin{tabular}{ccccccccccc}
\toprule
\multirow{2}{*}{{Methods}} & \multicolumn{2}{c}{{CASIA1.0+}} & \multicolumn{2}{c}{{Columbia}} & \multicolumn{2}{c}{{IMD2020}} & \multicolumn{2}{c}{{AutoSplice}} & \multicolumn{2}{c}{{Average}} \\ 
\cmidrule(lr){2-3} \cmidrule(lr){4-5} \cmidrule(lr){6-7} \cmidrule(lr){8-9} \cmidrule(lr){10-11}
& {ACC} & {F1} & {ACC} & {F1} & {ACC} & {F1} & {ACC} & {F1} & {ACC} & {F1} \\ 
\hline
w/o Extractor      & .615 & {.606} & .748 & {.727} & .594 & {.587} & {.761} & {.756} & {.680} & {.669} \\
w/o Prompt         & .789 & {.783} & .863 & {.851} & .707 & {\textbf{.715}} & {.751} & {.739} & {.778} & {.772} \\
w/o Mask Encoder   & .746 & {.738} & .834 & {.818} & .671 & {.657} & {\textbf{.791}} & {.752} & {.761} & {.741} \\
\hdashline
{with Qwen} & {.789} & {.776} & {.866} & {.852} & {.703} & {.712} & {.755} & {.741} & {.778} & {.770} \\
{with Intern} & {.757} & {.733} & {.899} & {.876} & {.711} & {.692} & {.739} & {.738} & {.776} & {.760} \\
% {with MiniGPT} & {.760} & {.763} & {.835} & {.822} & {.691} & {.666} & {.692} & {.683} & {.745} & {.734} \\
\hline
Ours & \textbf{.805} & {\textbf{.794}} & \textbf{.903} & {\textbf{.900}} & \textbf{.727} & {.709} & {.780} & {\textbf{.760}} & {\textbf{.804}} & {\textbf{.791}} \\
\bottomrule
\end{tabular}
    }
    \label{tab:ablation_detect}
\end{table*}

\begin{table*}[ht]
\centering
\vspace{-1em}
\caption{{Ablation results using different variants of our proposed ForgeryGPT on the image forgery localization task.}}
\resizebox{1\linewidth}{!}{
\begin{tabular}{ccccccccccccccccccc}
\toprule
\multirow{3}{*}{\textbf{Methods}} & \multicolumn{12}{c}{\textbf{Editing}} & \multicolumn{4}{c}{\textbf{DGM}} & \multicolumn{2}{c}{\multirow{2}{*}{\textbf{Average}}} \\ 
\cmidrule(lr){2-13} \cmidrule(lr){14-17}
                         & \multicolumn{2}{c}{CASIA1.0+} & \multicolumn{2}{c}{Columbia} & \multicolumn{2}{c}{NIST16} & \multicolumn{2}{c}{IMD2020} & \multicolumn{2}{c}{DSO-1} & \multicolumn{2}{c}{Korus} & \multicolumn{2}{c}{AutoSplice} & \multicolumn{2}{c}{OpenForensics} & \multicolumn{2}{c}{} \\ 
\cmidrule(lr){2-19}
                         & F1 & AUC & F1 & AUC & F1 & AUC & F1 & AUC & F1 & AUC & F1 & AUC & F1 & AUC & F1 & AUC & F1 & AUC \\ 
\hline
w/o Object & .511 & .847 & {.751} & {.882} & .509 & .832 & {.521} & {.891} & {.487} & {.851} & {.249} & {.716} & {.480} & {.861} & {.117} & {.691} & {.453} & {.821} \\
w/o Vocab & .423 & .797 & {.765} & {.902} & .404 & .737 & {.519} & {.902} & {.503} & {.872} & {.227} & {.700} & {.493} & {.844} & {.122} & {.661} & {.432} & {.802} \\
w/o Multi-scale & .542 & .852 & {.693} & {.876} & .512 & .839 & {.509} & {.910} & {.493} & {.847} & {.254} & {.729} & {.561} & {.869} & {.099} & {.659} & {.458} & {.823} \\
\hline
Ours & \textbf{.569} & \textbf{.876} & {\textbf{.773}} & {\textbf{.911}} & \textbf{.549} & \textbf{.883} & {\textbf{.530}} & {\textbf{.914}} & {\textbf{.506}} & {\textbf{.895}} & {\textbf{.258}} & {\textbf{.737}} & {\textbf{.570}} & {\textbf{.877}} & {\textbf{.138}} & {\textbf{.701}} & {\textbf{.536}} & {\textbf{.870}} \\
\bottomrule
\end{tabular}
}
\label{tab:ablation_localization}
\vspace{-1em}
\end{table*}

\begin{table}[!t]
\tiny
    \centering
    \caption{{Comparison Between LLaVA, GPT4o, and ForgeryGPT for Explanation Accuracy.}}
    \resizebox{0.8\linewidth}{!}{
   \begin{tabular}{c|ccc}
        \hline
        Metric     & LLaVA & GPT4o & Ours \\ \hline
        {CSS}        & {.687}  & {.699}  & {\textbf{.749}} \\
        {BLEU-4}     & {.193}  & {.229}  & {\textbf{.240}} \\
        ROUGE-L    & .218  & .244  & \textbf{.284} \\
        {SPICE}      & {.266}  & {.289}  & {\textbf{.309}} \\
        {Average}    & {.341}  & {.365}  & {\textbf{.395}} \\
        \hline
    \end{tabular}
    }
    \label{tab:rouge_scores}
\end{table}

% 这段全要高亮的用pdf高亮吧
\subsection{Image Forgery Localization}
% \subsubsection{Image Forgery Localization.} 
Following DiffForensics \cite{yu2024diffforensics}, we benchmark ForgeryGPT against a broad range of state-of-the-art methods under consistent experimental settings.
% Specifically, evaluations are conducted using the pre-trained models of ManTra-Net \cite{Mantrawu2019mantra} and MVSS-Net \cite{MVSS-Net2021image}.
Specifically, we evaluate representative methods including H-LSTM \cite{bappy2019hybrid}, HP-FCN \cite{li2019localization}, GSRNet \cite{zhou2020generate}, SPAN \cite{SPANhu2020span}, SATL-Net \cite{zhuo2022self}, CAT-Net \cite{kwon2021cat}, PSCCNet \cite{PSCCliu2022pscc}, HiFi-Net \cite{HiFi_IFDL}, and HiFi-Net++ \cite{guo2024language}, all trained using their official implementations on the same dataset. In addition, ManTra-Net \cite{Mantrawu2019mantra} and MVSS-Net \cite{MVSS-Net2021image} are evaluated directly using the pre-trained models provided by their respective authors.
We further include the latest LLM-based image forgery detection approaches, FAKESHIELD \cite{xu2024fakeshield} and SIDA \cite{huang2025sidasocialmediaimage}, which are trained following their official pipelines under identical data conditions.
Table~\ref{tab:localization} reports the pixel-level localization performance of all competing methods across eight benchmark datasets, using pixel-level AUC and F1-score as evaluation metrics. ForgeryGPT achieves the highest localization results on CASIA1.0+, NIST16, IMD2020, DSO-1, and Korus, and attains the second-best performance on AutoSplice and Columbia. Notably, it reaches an AUC of 91.4\% on IMD2020, a dataset consisting of real-world manipulated images, demonstrating its strong capability to detect subtle tampering traces while generalizing effectively to realistic scenarios.
These results highlight the effectiveness of ForgeryGPT, which leverages the FL-Expert to accurately capture diverse and fine-grained manipulation cues. Overall, ForgeryGPT achieves the highest average performance across all the datasets, underscoring its robustness and superiority in addressing the challenges posed by the advanced image forgery localization task.

\subsection{Image Forgery Detection}
We perform image-level classification experiments on datasets comprising both authentic and manipulated images. To ensure a fair comparison, all baseline methods are retrained under identical experimental settings. Table~\ref{tab:detection} summarizes the detection performance of the evaluated methods.
ForgeryGPT achieves the highest average accuracy and F1-score. While several baseline models are tailored for specific types of forgeries, for instance, HiFi-Net++ attains strong results on the CASIA1.0+ dataset, achieving an accuracy of 0.822 and an F1-score of 0.812. ForgeryGPT maintains consistently superior average performance across diverse manipulation techniques and domains.
These findings underscore the robustness of ForgeryGPT in handling varied forgery scenarios, demonstrating its effectiveness as a general-purpose and high-performing solution for real-world forgery detection.

% \begin{figure}[!t]
%     \centering
%     \includegraphics[width=0.9\linewidth]{figs/key-entity.pdf}
%     \caption{Evaluation of the values of the object-agnostic learnable word embedding size on the image forgery localization task (F1-score).}
%     \label{fig:key-entity}
% \end{figure}

\subsection{Robustness Evaluation}
To evaluate the robustness of our forgery localization model under real-world image degradations, we apply a series of controlled distortions to the original forged images from the NIST16 dataset, following the protocols outlined in \cite{Objectformerwang2022objectformer}. The applied distortions include multi-scale resizing (\textit{Resize}), Gaussian blurring with kernel size $k$ (\textit{GaussianBlur}), Gaussian noise with standard deviation $\sigma$ (\textit{GaussianNoise}), and JPEG compression with quality factor $q$ (\textit{JPEGCompress}).
We evaluate the forgery localization performance of our pre-trained model in terms of AUC scores and compare it against MVSS-Net, PSCCNet, and HiFi-Net under these distorted conditions. The results, summarized in Table~\ref{tab:robust}, indicate that our model consistently outperforms the baselines across all types of distortions. In particular, it exhibits a pronounced advantage under JPEG compression, a common scenario for images shared on social media, demonstrating its robustness and reliability in practical, real-world applications.

\subsection{Explainability Analysis}
% 可解释性的定性和定量分析(GPT4o模拟人工评测)
We evaluate the quality of explanations produced by ForgeryGPT with respect to both accuracy and persuasiveness. For this purpose, a testing set is constructed by randomly sampling 100 forged images from the IMD2020 dataset, followed by manual review in accordance with the previous procedure. The explanations generated by ForgeryGPT are then compared against the ground-truth descriptions of the forgeries in the testing set. The evaluation metrics encompass two categories: (1) Consistency metric, and (2) Human evaluation.

\subsubsection{Consistency Metric}    
To evaluate the overall quality and consistency of the explanations generated by ForgeryGPT, we adopt multiple complementary metrics, including ROUGE-L scores~\cite{lin2004rouge}, Cosine Semantic Similarity (CSS), BLEU-4 \cite{papineni2002bleu}, and SPICE \cite{anderson2016spice}. These metrics jointly provide a comprehensive assessment of explanation quality from both lexical and semantic perspectives.
We benchmark ForgeryGPT against strong baselines, including GPT-4o and LLaVA. As summarized in Table~\ref{tab:rouge_scores}, ForgeryGPT consistently attains the highest scores across all evaluation metrics. 
The superior ROUGE-L, BLEU-4, and CSS scores reflect its strong lexical alignment with reference explanations, while the elevated SPICE score highlights its enhanced semantic coherence and contextual relevance.
Overall, these results demonstrate that ForgeryGPT surpasses existing methods in generating accurate, coherent, and interpretable explanations for image forgeries.

% BLEU-4 \cite{papineni2002bleu}, and SPICE \cite{anderson2016spice}

\begin{figure}[!t]
    \centering
    \includegraphics[width=0.9\linewidth]{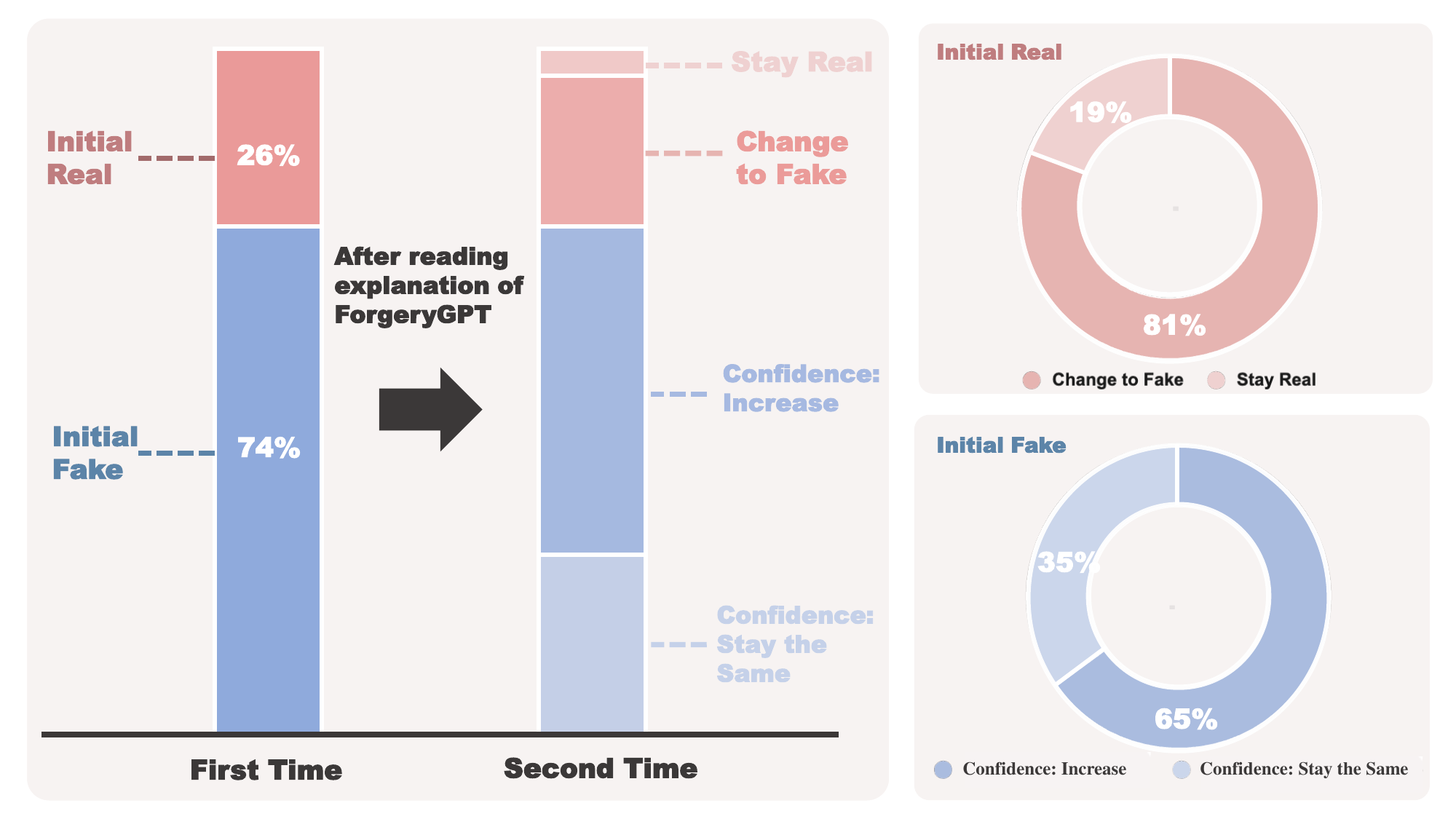}
    \caption{Results of human evaluation. The left side shows the initial distribution of test samples classified by the participants as either real (authentic) or fake (forgery), while the right side illustrates the changes in user judgment and confidence after reviewing the explanations generated by ForgeryGPT.}
    \vspace{-3mm}
    \label{fig:human_eval}
\end{figure}

\subsubsection{Human Evaluation} To further evaluate the practical impact of ForgeryGPT in exposing misinformation through its generated explanations, we conduct a human evaluation on these 100 forged images from the testing set. 
Five participants are asked to determine the authenticity of each image (\textit{i.e.}, real or fake) and to indicate any change in their confidence level (\textit{i.e.}, stay the same or increase), both before and after reviewing ForgeryGPT’s explanations. 
As illustrated in Figure~\ref{fig:human_eval}: 1) 74\% of the images are correctly recognized as forgeries by the participants, reflecting a baseline level of discernment in identifying manipulated content; 2) Among the images initially misclassified as real, 81\% are subsequently correctly identified as fake after consulting ForgeryGPT’s explanations, demonstrating the model’s strong persuasive capability; 3) for the images initially identified as fake, the explanations further enhance the participants’ confidence in their judgments by 65\%. Overall, these findings highlight ForgeryGPT’s capability to generate explanations that not only improve human understanding but also strengthen confidence and accuracy in detecting image forgeries.

\begin{figure}[!t]
    \centering
    \includegraphics[width=\linewidth]{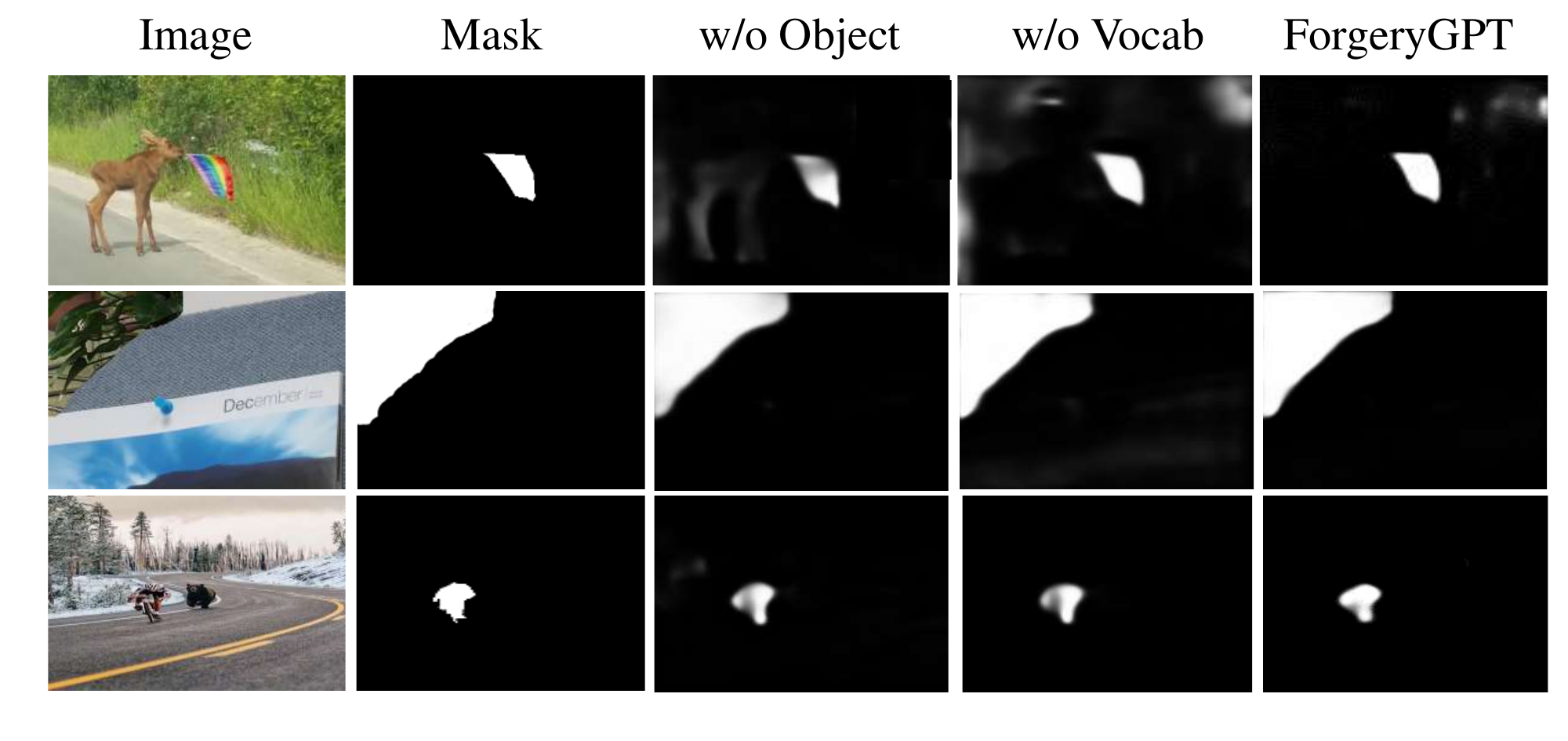}
    \caption{Visualizing the performance impact of ForgeryGPT across different variants.}
    \label{fig:visual_ablation}
\end{figure}

\begin{table}[!t]
  \centering
    \caption{The performance on NIST16 dataset under various distortions. AUC scores are reported (in $\%$), (Blur: GaussianBlur, Noise: GaussianNoise, Compress: JPEGCompress.)}
  \resizebox{\linewidth}{!}{
  \begin{tabular}{c|c c c c c}
    \toprule
    Distortion          & MVSS  & PSCC & HiFi & \multicolumn{2}{c}{Ours} \\
    \midrule
    no distortion       & 84.5 & 87.1   &  64.2  & \multicolumn{2}{c}{\textbf{87.6}}  \\ \hline
    Resize($0.78 \times$)        & 83.7 & 86.9  &  62.9   & \textbf{87.2} & { \textbf{$\downarrow$}0.4}\\ 
    Resize($0.25 \times$)        & 80.8 & 86.33    & 62.5    & \textbf{87.0}& { \textbf{$\downarrow$}0.6}\\ \hline
    Blur($k=3$)   & 82.9 & 86.1     & 63.1   & \textbf{86.2} & { \textbf{$\downarrow$}1.4}\\
    Blur($k=15$)  & 80.1 & 82.3    & 55.0    & \textbf{85.9} & { \textbf{$\downarrow$}1.7}\\ \hline
    Noise($\sigma=3$)  & 76.1 & 80.2   & 58.9     & \textbf{87.1} & { \textbf{$\downarrow$}0.5}\\
    Noise($\sigma=15$) & 68.8 & 79.2   & 56.4     & \textbf{86.8}  & { \textbf{$\downarrow$}0.8}\\ \hline
    Compress($q=100$) & 84.2 & 86.3   & 63.1     & \textbf{86.0} & { \textbf{$\downarrow$}1.6}\\ 
    Compress($q=50$)  & 82.8 & 86.2   & 63.0     & \textbf{85.9} & { \textbf{$\downarrow$}1.7}\\ 
    \bottomrule
  \end{tabular}}
  \label{tab:robust}
\end{table}

\begin{table}[!t]
\centering
\caption{Pixel-level AUC and classification accuracy (\%) on MVTec-AD and VisA benchmarks under few-shot settings. Only ForgeryGPT and AnomalyGPT results are shown.}
\resizebox{1\linewidth}{!}{
\begin{tabular}{l|l|cc|cc}
\toprule
\textbf{Setup} & \textbf{Method} & \multicolumn{2}{c|}{\textbf{MVTec-AD}} & \multicolumn{2}{c}{\textbf{VisA}} \\
 & & Pixel-AUC & Accuracy & Pixel-AUC & Accuracy \\
\midrule
\multirow{2}{*}{1-shot}
& ForgeryGPT & .751 & .669 & .681 & .604 \\
& AnomalyGPT \cite{gu2024anomalygpt} & .953 & .861 & .962 & .774 \\
\midrule
\multirow{2}{*}{2-shot}
& ForgeryGPT & .801 & 693 & .755 & .642 \\
& AnomalyGPT \cite{gu2024anomalygpt} & .956 & .848 & .964 & .775 \\
\midrule
\multirow{2}{*}{4-shot}
& ForgeryGPT & .790 & 677 & .743 & .633 \\
& AnomalyGPT \cite{gu2024anomalygpt} & .962 & .850 & .967 & .777 \\
\bottomrule
\end{tabular}
}
\label{tab:main_anomaly_fewshot}
\vspace{-3mm}
\end{table}

\begin{figure*}
    \centering
    \includegraphics[width=0.9\textwidth]{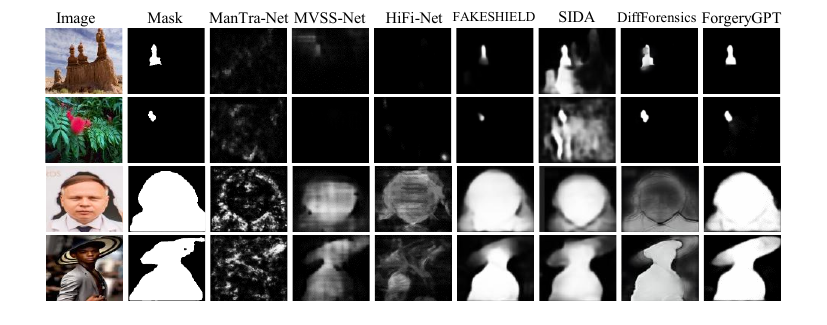}
    \caption{{Visualization of the predicted manipulation mask. From left to right, we show forged images, GT masks, and predictions of ManTraNet, MVSS-Net, HiFi-Net, FAKESHIELD, SIDA, DiffForensics, and ForgeryGPT.}}
    \label{fig:visual_all}
\end{figure*}

% \begin{figure*}[t]
%     \centering
%     \includegraphics[width=0.8\textwidth]{figs/clip-visual.pdf}
%     \caption{Perception capability of CLIP for the authenticity-forgery attributes of images. For each object prompt, we prepend a corresponding status modifier (e.g., ``\textit{a photo of a forged}'', ``\textit{a photo of an authentic}'') and compute the average score.}
%     \label{clip-visual}
% \end{figure*}

\begin{figure}
    \centering
    \includegraphics[width=0.8\linewidth]{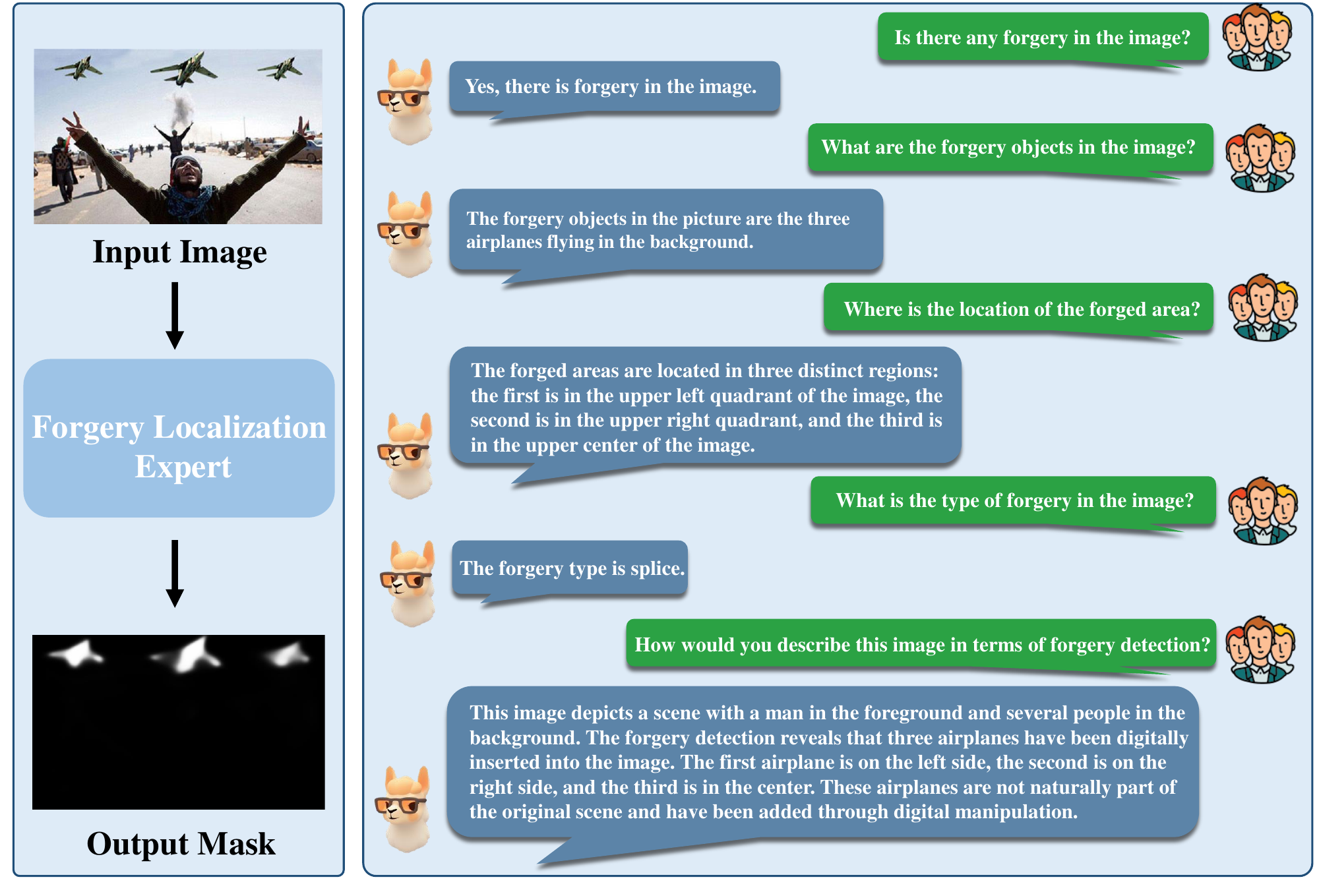}
    \caption{The multi-turn dialogue capability of ForgeryGPT on the IFDL task.}
    \vspace{-5mm}
    \label{fig:multi-turn-dialog}
\end{figure}

\subsection{Ablation Study}
\subsubsection{Analysis of Mask-Aware Forgery Extractor} We conduct ablation experiments to evaluate the effectiveness of the proposed Mask-Aware Forgery Extractor, as summarized in Table \ref{tab:ablation_detect}. \textit{w/o Extractor} denotes ForgeryGPT without using the mask token from the Mask-Aware Forgery Extractor for forgery detection, while \textit{w/o Prompt} indicates ForgeryGPT without using the learnable prompt token to bridge the image and mask modalities. 
\textit{w/o Mask Encoder} refers to ForgeryGPT without using the Mask Encoder to obtain a mask token (replacing it with the Image Encoder).
Experimental results reveal that the Mask-Aware Forgery Extractor significantly improves detection performance. These findings demonstrate that  
integrating fine-grained representations of localized mask regions effectively enriches forgery-related semantic cues, thereby enhancing the reasoning capability and accuracy of forgery detection.

\subsubsection{Analysis of FL-Expert}
% ``\textit{A photo of a forged person}''
The experimental results in Table \ref{tab:ablation_localization} demonstrate that the proposed FL-Expert plays a pivotal role in improving localization performance.
\textit{w/o Object} denotes FL-Expert without using the Object-agnostic Forgery Prompt (replacing with an ``\textit{object}'' word token). 
The observed improvement in performance indicates that the Object-agnostic Forgery Prompt automatically generates suitable object prompts, thereby facilitating accurate forgery localization.
In addition, \textit{w/o Vocab} refers to FL-Expert without using the Vocabulary-enhanced Vision Encoder. The corresponding results validate the advantage of introducing a new fine-grained visual vocabulary network, which strengthens the original vision network's representational capabilities and improves its sensitivity to subtle manipulation cues.
Furthermore, \textit{w/o Multi-scale} represents FL-Expert without using Multi-scale Fusion and Multi-layer Attention Fusion mechanisms to integrate multi-scale features (replacing them with the last layer features of ViT). 
The results indicate that extracting and aggregating information across multiple granularities contributes to the localization of image forgery.
% Finally, we analyze the influence of the object-agnostic word embedding size $m$. A smaller $m$ increases the risk of discarding relevant object-agnostic cues, while a larger $m$ heightens the risk of incorporating irrelevant noise. As illustrated in Figure~\ref{fig:key-entity}, $m = 12$ yields optimal localization performance.

\subsubsection{Analysis of MLLMs Backbone}
As shown in Table~\ref{tab:ablation_detect}, we evaluate the impact of different MLLMs backbones on ForgeryGPT, including Qwen2.5-VL-7B \cite{Qwen2.5-VL} and InternVL-2.5-8B \cite{chen2024internvl}. The variants ``\textit{with Qwen}'' and ``\textit{with Intern}'' denote configurations adopting these models as the MLLMs backbone, respectively. All variants are trained following the same three-stage pipeline and dataset as the ForgeryGPT.
% Among these alternatives, ForgeryGPT with Vicuna and CLIP achieves the best overall performance, attributed to the comprehensive pretraining and instruction-tuning data available from LLaVA \cite{llava}, which endows the model with strong visual grounding and reasoning capabilities.
% In contrast, models such as Qwen2.5-VL-7B and InternVL-2.5-8B are limited by data accessibility, though Qwen remains competitive owing to its robust architecture. These results highlight the impact of both backbone design and training data, and validate the modular flexibility of the proposed framework.
% Among these alternatives, ForgeryGPT achieves the best overall performance, attributed to the 创新性的针对IFDL任务所设计的一体化大模型架构，从而实现了对伪造掩码信息的精确挖掘，以及对篡改痕迹的像素级理解。相比之下，诸如 Qwen2.5-VL-7B 和 InternVL-2.5-8B 等模型虽然受限于数据的可获取性，但 Qwen 凭借其稳健的架构依然保持了竞争力。这些结果凸显了骨干网络设计（Backbone Design）和训练数据两者的重要影响，并验证了所提出框架的模块化灵活性。
% 相比之下，采用 Qwen2.5-VL-7B 和 InternVL-2.5-8B 作为 MLLM 主干网络会导致性能下降，其主要原因在于这些模型所依赖的预训练数据在可获取性和覆盖度方面存在局限。尽管如此，Qwen 依托其稳健的架构设计仍保持了相对竞争力。上述结果凸显了主干网络设计与预训练数据质量两者的共同重要性，并进一步验证了 ForenAgent 所采用 backbone 的有效性。
In contrast, using Qwen2.5-VL-7B and InternVL-2.5-8B as MLLM backbones results in a performance drop, largely due to limitations in the accessibility and coverage of their corresponding pre-training data. Nevertheless, Qwen remains relatively competitive thanks to its robust architectural design. These findings highlight the intertwined importance of backbone architecture and pre-training data quality, and further validate the effectiveness of the backbone adopted in our ForgeryGPT framework.

% \begin{figure*}
%     \centering
%     \includegraphics[width=1\textwidth]{figs/visual.pdf}
%     \caption{{Illustration of three detection cases using ForgeryGPT. Correctly predicted content is marked in green, while incorrect outputs are highlighted in red.}}
%     \label{fig:new_visual_fail}
% \end{figure*}

\subsection{Visualization Results}
% (1) 定位效果可视化 (2) 可解释性case可视化 (3) 在其他领域效果的可视化
\subsubsection{Qualitative Results} 
Figure~\ref{fig:visual_all} presents comprehensive qualitative comparisons across a wide range of datasets, including CASIA1.0+, Columbia, NIST16, IMD2020, DSO-1, Korus, and AutoSplice, encompassing both general object and face forgery scenarios. For each example, from left to right, we display the forged image, ground-truth (GT) mask, and predicted manipulation masks from ManTraNet, MVSS-Net, HiFi-Net, DiffForensics, as well as the recent LLM-based approaches FAKESHIELD~\cite{xu2024fakeshield} and SIDA~\cite{huang2025sidasocialmediaimage}, followed by our ForgeryGPT.
The visualization results clearly indicate that ForgeryGPT consistently achieves more precise localization of tampered regions, yielding sharper and better-defined boundaries.
This improvement can be largely attributed to the integration of the FL-Expert module, which allows the model to effectively discern subtle inconsistencies and fine-grained manipulations within complex visual scenes.

\subsubsection{Visualization of Object-agnostic Forgery Prompt} To investigate the impact of the Object-agnostic Forgery Prompt, we visualize the feature changes with and without prompt learning in Figure~\ref{fig:visual_ablation}. \textit{w/o Object} represents FL-Expert without using the Object-agnostic Forgery Prompt (replacing with an ``\textit{object}'' word token). It can be observed that our design substantially enhances the accuracy of forgery localization, highlighting its effectiveness in guiding the model to focus on relevant manipulated regions.
% Without OTPL, the network tends to make false judgments about objects that are similar to the forged ones.

\subsubsection{Visualization of Vocabulary-enhanced Vision Encoder} We further investigate the feature changes with and without the Vocabulary-enhanced Vision Encoder in Figure~\ref{fig:visual_ablation}. \textit{w/o Vocab} represents FL-Expert without using the Vocabulary-enhanced Vision Encoder. The results clearly indicate that the Vocabulary-enhanced Vision Encoder promotes more effective learning of forgery-related features and yields more precise contours for the manipulated regions.

\subsubsection{Visualization of Multi-turn Dialog} 
Figure~\ref{fig:multi-turn-dialog} illustrates the multi-turn dialogue capabilities of ForgeryGPT. It demonstrates that ForgeryGPT can accurately evaluate the authenticity of images and generate corresponding forgery masks.
Moreover, it is capable of engaging in interactive dialogue to precisely identify forged objects, specify forgery regions, determine forgery type, and provide a summary of the forgery details. This interactive capability highlights ForgeryGPT’s ability to combine precise forgery detection with natural language reasoning, providing users with a comprehensive and interpretable understanding of image manipulations.

\subsection{Generalization to Other Domains} 
To evaluate cross-domain generalization, we benchmark ForgeryGPT on industrial anomaly localization using the VisA \cite{zouvisa} and MVTec \cite{bergmann2019mvtec} datasets, following the AnomalyGPT \cite{gu2024anomalygpt} protocol. 
As summarized in Table~\ref{tab:main_anomaly_fewshot}, ForgeryGPT achieves competitive results in both pixel-level AUC and image-level accuracy. The model demonstrates robust zero-shot transferability, underscoring its capacity for diverse visual reasoning tasks that extend beyond its original training scope.

% \subsubsection{Visualization of Failure Cases}
% To thoroughly investigate the limitations of ForgeryGPT, we present three representative cases in Figure~\ref{fig:new_visual_fail}. {Case (a) shows that without the forgery mask, the model fails to detect manipulation; however, once aided by the FL-Expert-generated mask, it makes a correct prediction with improved interpretability, demonstrating the importance of the localization module. Case (b) reveals that an inaccurate mask can mislead the LLM's explanation, even if the image is correctly classified as manipulated, highlighting the dependency of interpretability on accurate forgery mask grounding, which can be mitigated by providing more precise forgery masks. Case (c) demonstrates the model’s robustness, where it correctly detects the forgery despite receiving a mask with false positive regions, owing to the mask encoder’s design that compresses mask cues while preserving comprehensive visual features necessary for reasoning.}

\section{Conclusion}
% In this work, we introduce ForgeryGPT, a novel framework that significantly advances the IFDL task by leveraging high-order forensics knowledge correlations from diverse linguistic feature spaces, while offering explainable generation and interactive dialogue capabilities. Through the integration of the Mask-Aware Forgery Extractor and the three-stage training process: Image-Text Alignment Pre-training, Mask-Text Alignment Pre-training, and Task-Specific Instruction Tuning—ForgeryGPT addresses critical limitations of existing methods, such as the reliance on manual thresholds and the lack of interpretability. Our comprehensive experiments demonstrate that ForgeryGPT achieves superior performance as compared to both the MLLM GPT-4o and state-of-the-art IFDL methods. Additionally, ForgeryGPT excels at providing detailed, convincing explanations, as confirmed through quantitative analysis and human evaluation. These results highlight ForgeryGPT’s potential as a robust solution for fine-grained forgery detection and localization, contributing significantly to the advancement of explainable AI in this domain.
In this work, we introduce ForgeryGPT, a novel framework that advances the IFDL task by leveraging high-order forensic knowledge correlations from diverse linguistic feature spaces, while providing explainable generation and interactive dialogue capabilities. 
% Through the integration of the Mask-Aware Forgery Extractor and a carefully designed three-stage training pipeline—Image-Text Alignment Pre-training, Mask-Text Alignment Pre-training, and Task-Specific Instruction Tuning—ForgeryGPT addresses critical limitations of existing approaches, including reliance on manual thresholds and lack of interpretability.
Extensive experiments demonstrate that ForgeryGPT consistently outperforms both the MLLM GPT-4o and state-of-the-art IFDL baselines across multiple benchmarks. In addition, the model excels at generating detailed and trustworthy explanations, as confirmed through both quantitative metrics and human evaluation, highlighting its ability to integrate vision-language understanding with forensic reasoning.
Despite these achievements, we acknowledge that the incorporation of large language models introduces inherent risks of hallucination, particularly when the visual evidence is ambiguous or the prompt context is limited. Addressing such challenges remains an important avenue for future research. Potential directions include the development of more robust pretraining paradigms, enhanced grounding mechanisms, and the utilization of larger and more diverse multimodal datasets to further bolster the model’s reliability and generalization.
Overall, ForgeryGPT marks a promising stride toward building accurate, interpretable, and user-aligned systems for fine-grained image forgery detection and localization, contributing to the broader pursuit of trustworthy and explainable AI.

\bibliographystyle{IEEEtran}
\bibliography{Main}

% \newpage

% \appendix

% \section{}

\vfill

\end{document}